\documentclass{article}

\usepackage[preprint]{nips}

\usepackage[utf8]{inputenc} % allow utf-8 input
\usepackage[T1]{fontenc}    % use 8-bit T1 fonts
\usepackage{amsfonts}       % blackboard math symbols
\usepackage{nicefrac}       % compact symbols for 1/2, etc.
\usepackage{microtype}      % microtypography

\usepackage{graphicx}
\usepackage{subcaption}
\usepackage{amsmath}
\usepackage{amssymb}

\usepackage{algorithmic}
\usepackage{algorithm}
\usepackage{authblk}

\usepackage{amsmath}
\usepackage{relsize}
\usepackage{amssymb}
\usepackage{url,hyperref, nameref}
\usepackage{amsthm}
\newtheorem{theorem}{Theorem}

\newtheorem{lemma}[theorem]{Lemma}
\theoremstyle{definition}
\newtheorem{proposition}[theorem]{Proposition}
\usepackage{algorithm}
\usepackage{algorithmic}
\title{Cross Learning in Deep Q-Networks}

\author{\textbf{Xing Wang}, \quad
\textbf{Alexander Vinel}\thanks{Corresponding to: alexander.vinel@auburn.edu} }

\affil{Department of Industrial Engineering\\ Auburn University\\ Auburn, AL, 36849, USA}

\begin{document}
\maketitle

\begin{abstract}
In this work, we propose a novel \textit{cross Q-learning} algorithm, aim at alleviating the well-known overestimation problem in value-based reinforcement learning methods, particularly in the deep Q-networks where the overestimation is exaggerated by function approximation errors. 
Our algorithm builds on double Q-learning, by maintaining a set of parallel models and estimate the Q-value based on a randomly selected network, which leads to reduced overestimation bias as well as the variance. 
We provide empirical evidence on the advantages of our method by evaluating on some benchmark environment, the experimental results demonstrate significant improvement of performance in reducing the overestimation bias and stabilizing the training, further leading to better derived policies.
\end{abstract}

% \begin{keyword}
% %% keywords here, in the form: keyword \sep keyword
% \sep Reinforcement learning
% \sep Multivariate forecasting
% \sep Multi-step forecasting
% \sep Deep learning
% \sep Convolutional neural network
% % \sep Embedding
% \end{keyword}

%% main text

\section{Introduction}
\label{sec:cq_intro}
Overestimation has been identified as one of the most severe problems in value-based reinforcement learning (RL) algorithms such as Q-learning \cite{thrun1993issues}, where the maximization of value estimates induces a consistent positive bias, and the error of the estimates is accumulated by the nature of temporal difference (TD) learning. In the function approximation setting such as deep Q-networks (DQN), the issue of value overestimation is more severe, given the noise induced by the inaccuracy of the approximation. 
As a result, learning DQN tends to have instability and variability for estimated Q-values, the derived policies accroding to the overestimated Q-values tend to be not optimal and often diverge. 

To overcome this issue, double Q-learning \cite{ddqn} has become a standard approach for training DQNs. The main purpose of double Q-learning is to avoid the overestimation problem for the target Q-value, by introducing negative bias from the double estimates. The usual way to realize it in DQN is to maintain a target network which is a copy of the policy DQN which is either frozen for a period of time, or softly updated with exponential moving average. The target network then is used to estimate the TD target. 
This may alleviate the issue, however, double DQN still often suffer from overestimation in practice, partially because the policy and target estimates of $Q$-values are usually too similar, while the noise from high variance is propagated through the network and occasional large reward can produce great overestimation in the future. 
Another approach sometimes proposed is to impose a bias-correction term on the estimates for Q-learning \cite{lee2013bias}, however, the error correction term is complicated to derive for deep networks, in which the finiteness of state space is no longer true.
A more recent modification over double DQN favors underestimation and clips the Q-value estimates \cite{fujimoto2018addressing}, that is, always chooses the minimum of the estimated targets over the two networks. The clipped double Q-learning is used on the critics in actor-critic methods for the deterministic policy gradient, which is referred to as TD3 (twin delayed deep deterministic policy gradient) and has shown state-of-the-art results on multiple tasks.
However, the intentionally engineered underestimation lacks of rigorous theoretical guide, in addition, it may  induce bias in the other direction, e.g., the underestimation can also accumulate through TD learning and derive suboptimal policies. Further, excessive underestimation can naturally lead to slower convergence.

Another direction to alleviate overestimation is through reducing the variance during training. For example, \cite{anschel2017averaged} uses the average of the learned estimated Q-values from multiple networks, which is designed to help  reduce the target approximation error.
There also exist various variance reduction techniques \cite{johnson2013accelerating, defazio2014saga, schmidt2017minimizing, allen2017katyusha} that focus on the general non-convex optimization procedure for accelerating the stochastic gradient descent, or their direct application on DQNs \cite{chen2019q}, in which the agent could obtain smaller approximated gradient errors. Reducing the variance can effectively stabilize the DQN training procedure, and overestimation alleviation can be seen as a by-product. However, these are indirect methods for overestimation control, and the positive bias due to the max operator in TD update are not taken care of.

To address these concerns, we propose a cross DQN algorithm, which can be seen as a direct extension of an earlier variant of double DQN, but can be more flexible. In cross DQN, we maintain more than two networks, and update them  one at a time based on the estimation from another randomly selected  one. %This has similar theoretical foundation with the well-known bagging (bootstrap aggregating) ensemble in general machine learning.
%i.e., we have a bag of $K$ networks, update can be realized by bootstrapping samples.%, and action can be chosen either by aggregating or ensemble (voting).
%If we go back and read the original paper [\cite{doubleq}], the author provided theoretical analysis on randomly choosing one Q-value to update based on the other Q, both are updated along time instead of being frozen for a period as a target. This can be extended to 
%In order to  illustrate the method, we highlight its comparison with existing work that maintain multiple networks. 
As  mentioned above, the averaged DQN \cite{anschel2017averaged} calculates the average of $K$ estimated Q-values, with the primary purpose of the overall variance reduction. For all $K$ networks, each step of TD updates as well as action selections are based on combining the $K$ estimates.  Consequently, the networks are tangled together and cannot be implemented with a parallel simulation. 
In bootstrapped DQN \cite{osband2016bdqn}, one of the $K$ networks (or heads) is bootstrapped for each action selection step during training, aiming at encouraging exploration early on. Thus the simulation is not independent among networks,
while the TD updates are totally independent within each of the networks, by using its own estimation of Q-values as in standard (double) DQN.
\cite{chen2018ensemble} investigates more general applications of traditional ensemble reinforcement learning on policies, i.e., majority voting, rank voting, Boltzmann addition, etc. to combine the different policies derived from multiple networks, by which they called the target ensembles, in addition to the averaged DQN which they called the temporal ensemble. %\red{what was their conclusion? Why do you mention this here?}
All of the above-mentioned work that maintain multiple networks have achieved better performance by addressing different issues through some particular settings. 
Our method focuses on the variation of TD updates, in which the target Q-values are estimated with a bootstrapped network for calculating the gradients, with the direct goal of reducing overestimation. Each of the $K$ networks would perform its own TD updates, while maintaining flexibility in action selections:
the networks can either interact with the environment independently, or through any other  ensemble strategy. The detailed implementation options would be discussed in Section \ref{sec:cq_cross}.

%We would also like to briefly discuss the linkage and motivation of our proposed method in statistical learning theory.
In supervised learning, ensemble strategies such as bagging, boosting, stacking, and hierarchical mixture of experts, etc. are commonly applied to achieve better performance, by simultaneously learning and combining multiple models. 
All of  the abovementioned algorithms that maintain multiple models, including ours, can be seen  as  special cases of general ensemble DQNs.
But our method has a deeper root in resampling and model selection. By bootstrapping another model to assess the values of current model, we introduce model bias for in-sample estimations, but reduce the variance of out-of-sample estimations (i.e., the squares of out-of-sample bias), in other words, the trained model can generalize better and alleviate overfitting. For squared errors, this can be expressed as the well-known bia-variance trade-off: 
$\textrm{MSE} =  \textrm{Irreducible Error}^2 + \textrm{Bias}^2 + \textrm{Variance}.$
In value-based reinforcement learning, the model easily overfits due to overestimation (which is caused by the max operator) during learning. Cross Q-learning introduces underestimation bias, and further reduces the variance, thus improves the generalization of the trained model. %\red{I can't make sense of what you are saying in this paragraph}

Like in \cite{fujimoto2018addressing}, our work can be naturally extended to the state of the art actor-critic methods in continuous action space, such as the deep deterministic policy gradient \cite{lillicrap2015continuous}, in which the critic network(s) are learned to give an estimate of the Q-value for the actor network to update its gradient and derive policies. Usually multiple critic networks are applied, however, rather than accumulating their learned gradients (either synchronously or asynchronously \cite{mnih2016asynchronous}) and optionally sharing network layers, no other information is shared among the critics. The extension of our method allows the critics to share their value estimates and utilize that of others, which leads to more accurate estimation of each critics, thus can improve the performance of these models. Similar to these actor-critic algorithms, our work can be implemented for parallel training easily, and the exchange of information among networks could take place either synchronously or asynchronously like the accumulation of gradients, as there is always tradeoff between synchronous and asynchronous update. 

%The above thinking procedure reminds us the recent development of Actor-Critic (AC) methods since 2016, all have used multiple workers (since there must be at least two different networks, one for actor and one for critic. Usually there is a single policy network, and multiple behavior networks to simulate and gather learning gradients. The recent work all focus on policy gradient actor, but the critics are still based on Q-value or its variant, i.e., advantage), from asynchronous to synchronous version.
% And our cross-learning idea can be easily extended to AC. None of the state of art work on ACs has taken advantage of sharing information from other networks while using multiple users, other than simply aggregating their learning gradients. 

The rest of this paper is organized as follows. In Section \ref{sec:cq_bg}, we resume the basics of value-based RL, and go through some recent related research. In Section \ref{sec:cq_est}, we formally define the estimators for the maximum expected values, along with their theoretical properties. The convergence of our cross estimator is shown in Section \ref{sec:cq_conv}. %In Section \ref{sec:gen}, we propose to generalize cross DQN and provide several variants, more importantly, we provide an interesting point of view by giving an analogy with supervised learning. 
Section \ref{sec:cq_cross} illustrates our cross DQN algorithm directly derived from the double DQN in details. 
We show some empirical results in Section \ref{sec:cq_res}. Finally, Section \ref{sec:cq_conc} draws conclusions and discusses future work.

\section{Background}
\label{sec:cq_bg}

\subsection{Value-based Reinforcement Learning}
% \subsection{Markov Decision Processes and Q-learning}
A natural abstraction for many sequential decision-making problems is to model the system as a \textit{Markov Decision Process} (MDP) \cite{puterman1994mdp}, in which the agent interacts with the environment over a sequence of discrete time steps. 
It is often represented as a 5-tuple:  $M=<\mathcal{S}, \mathcal{A}, T, R, \gamma>$, where $\mathcal{S}$ is a set of \textit{states}; $\mathcal{A}$ is a set of \textit{actions} that can be taken; $T: \mathcal{S}\times \mathcal{A} \mapsto \mathcal{P_S}$ is the \textit{transition function} such that $\int_{s'\in\mathcal{S}} T(s'|s, a) = 1$, which denotes the (stationary) probability distribution over $\mathcal{S}$ of reaching a new state $s'$, after taking action $a$ in state $s$; $R$ is the reward function, which can take the form of either 
$R:\mathcal{S} \mapsto \mathbb{R}$, $R:\mathcal{S}\times \mathcal{A} \mapsto \mathbb{R}$, or $R: \mathcal{S}\times \mathcal{A}\times \mathcal{S} \mapsto \mathbb{R}$; and $\gamma \in [0, 1)$ is the \textit{discount factor}.

A policy $\pi: \mathcal{S} \mapsto \mathcal{P_A}$ defines the conditional probability distribution of choosing each action while in state $s$. For an MDP, once a stationary policy is fixed, the distribution of the reward sequence is then determined. Thus to evaluate a policy $\pi$, it is natural to define the \textit{action value function under $\pi$} as the expected cumulative discounted reward by taking action $a$ starting from state $s$ and following $\pi$ thereafter:
\begin{align}
  Q^\pi (s, a) \equiv \mathbb{E}_\pi \Big[\sum_{\tau=0}^\infty \gamma^\tau R_{t+\tau} |S_t = s, A_t = a \Big]  
     = R(s, a) + \gamma \int_{s'} T(s'|s, a) Q^\pi (s', \pi(s')). 
\end{align}

The goal of solving an MDP is to find an \textit{optimal policy} $\pi^\ast$ that maximizes the expected cumulative discounted reward in all states. The corresponding \textit{optimal} action values satisfy $Q^{\ast} (s, a) = \max_\pi Q^\pi (s, a)$, and Banach's fixed-point theorem ensures the existence and uniqueness of the fixed-point solution of \textit{Bellman optimality equations} \textbf{\cite{puterman1994mdp}}: 
\begin{equation}
Q^\ast (s, a) = R(s, a) + \gamma \int_{s'} T(s'|s, a) \max_{a'} Q^\ast (s', a') 
\label{eqn:bellman}
\end{equation}
from which we can derive a deterministic optimal policy by being greedy with respect to $Q^\ast$, i.e., $\pi^\ast = \textrm{argmax}_{a\in \mathcal{A}} Q^\ast (s, a)$. 
% Given that the transition and reward functions are known and correct, solving the MDP is a \textit{planning} problem using \textit{dynamic programming} approaches. 
%Equantion (\ref{eqn:bellman}) provides a recursive form that can be solved via \textit{value iteration, policy iteration,} or \textit{linear programming}, all guarantee convergence %in polynomial time and allow us to obtain the optimal policy for the MDP.

In reinforcement learning problems, the agent must interact with the environment to \textit{learn} the information about the transition and reward functions, meanwhile trying to produce an optimal policy. While interacting with the environment, at each time step $t$, the agent senses some representation of current state $s$, selects an action $a$, then receives an immediate reward $r$ from the environment and finds itself in a new state $s'$. The \textit{experience tuple} $<s, a, r, s'>$ summarizes the observed transition for a single step. Based on the experiences through interacting with the environment, the agent can either learn the MDP model first by approximating the transition probabilities and reward functions, and then plan in the MDP to obtain an optimal policy (this is called the \textit{model-based} approach in reinforcement learning); or without learning the model, directly learn the optimal value functions and upon which the optimal policy is derived (this is called the \textit{model-free} approach). 

%We use Q-learning with function approximation in this paper. 
As a \textit{model-free} approach, Q-learning \cite{watkins1992q} updates one-step bootstrapped estimation of Q-values from the experience samples over time steps.  
The update rule upon observing $<s, a, r, s'>$ is
\begin{equation}
    Q(s, a) \leftarrow Q(s, a) + \alpha \big( r + \gamma \max_{a'} Q(s', a') - Q(s, a) \big)
\end{equation}
in which $\alpha$ is the learning rate,
$r+ \max_{a'} Q(s', a')$ serves as the update target of the Q-value, which can be seen as a sample of the expected value of one-step look-ahead estimation for state-action pair $(s, a)$, based on the the maximum estimated value over next state $s'$, 
and the last term $Q(s, a)$ is simply the current estimation. 
The difference $\delta = r + \gamma \max_{a'} Q(s', a') - Q(s, a)$ is referred to as temporal difference (TD) error, or Bellman error. 
Note that one can bootstrap more than one step when estimating the target, often by using the \textit{eligibility trace} as in $TD(\lambda)$ \cite{sutton1988td}.
Q-learning is guaranteed to converge to the optimal values in probability as long as 
each action is executed in each state infinitely often, $s'$ is sampled following the distribution $T(s, a, s')$, $r$ is sampled with mean $R(s, a)$, variance is bounded and given appropriately decaying $\alpha$. 

\subsection{DQN and Double DQN}

For environments with large state spaces, the Q-values are often represented by a function of state-action pairs rather than the tabular form, i.e., $Q_\theta(s, a) = f(s, a|\theta)$, where $\theta$ is a parameter vector. 
We consider Q-learning with function approximation in this paper. 
To update parameter vector $\theta$, first-order gradient methods are usually applied to minimize the mean squared error (MSE) loss:
$
    \theta \leftarrow \theta + \alpha \delta \nabla_{\theta} Q_\theta.
$
However, with function approximation, the convergence guarantee can no longer be established in general. Neural networks, while attractive as a powerful function approximator, were well known to be unstable and even to diverge when applied for reinforcement learning until deep Q-network (DQN) \cite{dqn} was introduced to show great success, in which several important modifications were made. 
\textit{Experience replay} \cite{lin1992experiencereplay} was used to address the non-stationary data problem, by storing and mixing the samples (i.e., experiences) into a replay memory for the updates. During training a batch of experiences is randomly sampled each time and the gradient descent is performed on the sampled batch. This way the temporal correlations could be alleviated. In addition, a separate \textit{target network}, which is  a copy of the learned network parameters ($\theta$) is employed. This copy is frozen for a period of time and is only updated periodically (denoted as $\theta^-$), and is applied to calculate the TD error, with the aim of improving stability. 

A variety of extensions and generalizations have been proposed and shown successes in the literature.
Overestimation due to the max operator in Q-learning may significantly hurt the performance. To reduce the overestimation error, double DQN (DDQN) \cite{ddqn} decouples the action selection from estimation of the target, that is, choosing the maximizing action according to the original network ($Q_\theta$), and evaluate the current value using the other one ($Q_{\theta^-}$ from the target network), i.e., 
$
Q_\theta (s, a) \leftarrow r + \gamma Q_{\theta^-} (s', \textrm{arg}\max_a Q_\theta (s', a)).
$
The procedures of double DQN is shown in Algorithm \ref{alg:ddqn}.

\begin{algorithm}[htb!]
\caption{Double DQN}
\label{alg:ddqn}
\begin{algorithmic}[1]
\footnotesize
\STATE Initialize policy network $Q_\theta$ and target network $Q_{\theta_-}$ with random parameters.
\STATE Initialize replay buffer $\mathcal{B}$.
% \REPEAT[for each episode]
\FOR{each episode \textbf{until} end of learning}
    \STATE Initialize state $s$
    % \STATE Randomly pick a network $Q^k$ to act, where $k\in \{1, \cdots, K\}$.
    % \REPEAT [for each step in an episode]
    % \For{each step in an episode}
    \FOR{step $t=1, \cdots$ \textbf{until} $s$ is terminal state of an episode} 
        \STATE Select action $a_t = \textrm{argmax}_{a} Q_\theta (s, a)$ with exploration
        %(with or without exploration?)
        % \STATE Select action  $a = \textrm{MajorityVote}\{ \textrm{argmax}_a Q_k(s, a) \}_{k=1}^K$  with exploration 
        \STATE Take action $a_t$, observe reward $r$ and next state $s'$
        \STATE Store experience tuple $<s, a_t, r, s'>$ into $\mathcal{B}$
        \STATE Sample a mini-batch of experiences from $\mathcal{B}$.
        % \STATE Randomly pick another network $Q^j$ for obtaining estimated target values.
        \FOR{all sampled experience in the mini-batch} 
            \STATE To train network $Q_{\theta}$, compute $a' = \textrm{argmax}_{a} Q_\theta (s', a)$
            \STATE Estimate TD target with target network $y = r + Q_{\theta_-}(s', a')$
            % \STATE Compute the TD error $\delta = \hat{Q} -Q^k(s, a|\theta^k)$, in which the target value is estimated using $Q^j$.
            \STATE Backpropagate TD error $\delta = y -Q_\theta(s, a_t)$ through $Q^k$, update $\theta$ with learning rate $\alpha$
        \ENDFOR
        \STATE $s\leftarrow s'$
        \STATE Update target network $\theta_- \leftarrow \theta$ in a fixed frequency
    \ENDFOR
    % \UNTIL{$s$ is terminal state}
\ENDFOR
    % \UNTIL{end of learning}
\end{algorithmic}
\end{algorithm}

\subsection{Dueling DQN}
\cite{wang2015dueling} proposed the dueling network architecture, in which lower layers of a deep neural network are shared and followed by two streams of fully-connected layers, that are used to represent two separate estimators, 
one for the state value function $V(s)$ and the other for the associated state-dependent action advantage function $A(s, a)$. The two outputs are then combined to estimate the action value $Q(s, a)$:
\begin{equation}
Q(s, a) = V(s) + A(s, a) - \frac{1}{|\mathcal{A}|} \sum_{a'} A(s, a')
\label{eqn:dueling}
\end{equation}
Note here the average of advantage values across all possible actions are used to achieve better stability, instead of the max operator in the other form proposed in \cite{wang2015dueling}, i.e., 
\begin{equation}
Q(s, a) = V(s) + A(s, a) - \max_{a'} A(s, a')
\label{eqn:dueling2}
\end{equation}
The dueling factoring often leads to faster convergence and better policy evaluation, especially in the presence of similar-valued actions. 
The deployment of advantage values is more robust to noise, since it emphasizes the gaps between $Q$-values of different actions given the same state, which are usually tiny thus small amount of noise may results in reordering of actions. In addition, the subtraction of an action-irrelevant baseline in Equation (\ref{eqn:dueling}) also effectively reduces variance, which helps stabilize learning and thus is more often used. The shared feature learning module also generalizes learning across actions, in which more frequent updating of the value stream $V$ leads to more efficient learning of state values, contrasts with that in DQNs of a single stream output, only one of the action values is updated while other action values remain untouched.

\subsection{Bootstrapped DQN}
The main purpose of Bootstrapped DQN \cite{osband2016bdqn} is to provide efficient ``deep'' exploration inspired by \textit{Thompson sampling} or as \textit{probability matching} in Bayesian reinforcement learning \cite{strens2000bayesian}, but instead of maintaining a distribution over possible values and intractable exact posterior update, it takes a single sample from the posterior. Bootstrapped DQN maintains a $Q$-ensemble, represented by a multi-head deep neural network in order to parameterize a set of $K\in \mathbb{N}_+$ different $Q$-value functions. The lower layers are shared by the $K$ ``heads'', and each head represents an independent estimate of the action value $Q^k(s, a | \theta^k)$. For each episode at training, Bootstrapped DQN picks a single head uniformly at random, and follows the greedy policy with respect to the selected $Q$-value estimates, i.e., $a_t = \textrm{argmax}_a Q^k(s_t, a)$, until the end of the episode. 

Bootstrapped DQN diversifies the $Q$-estimates and improves exploration through independent initialization of the $K$ heads as well as the fact that each head is trained with different experience samples. The $K$ heads can be trained together with the help of so-called bootstrap mask $m_k^\tau$, which decides whether the $k$-th head should be trained, i.e., the transition experience $\tau$ updates $Q_k$ only if $m^k_\tau$ is nonzero. In addition, bootstrapped DQN adapts double DQN in order to avoid overestimation, i.e., the the estimates of TD targets are calculated using the target network $Q_{\theta^k_-}$. The loss backpropagated to $k$-the head is then
\begin{equation}
    L(\theta^k) = \mathbb{E}_\tau [ m^k_\tau ( r + \gamma Q^k(s', a' | \theta^k_-) - Q^k(s, a | \theta^k))^2 ] \textrm{ where } a' = \textrm{argmax}_{a} Q^k (s', a | \theta^k)
    \label{eqn:bdqn}
\end{equation}
Note the gradients should be further aggregated and normalized for updating the lower layers of the network.

\section{Cross DQN}
\label{sec:cq_cross}
In this section, we elaborate our proposed cross Q-learning method and its variants. Cross DQN serves as an extension to the double DQN algorithm \cite{ddqn}, which has been used as the default setting for most state-of-art DQN training. 

Double DQN was proposed in the aim of reducing overestimation bias, in which the target network simply is a delayed-updated copy of the current network. Note that the original vanilla DQN also uses two networks, the purpose of periodic frozen and update of the target network is to stabilize learning. Specifically, in vanilla DQN, the target network is used to evaluate both the action and the value, i.e.,
\begin{equation}
    y \leftarrow r + \gamma Q_{\theta'} (s', a'_\ast) \qquad \textrm{ where } a'_\ast = \mathrm{argmax}_{a'} Q_{\theta'} (s', a')  
\end{equation}

On the other hand, in double DQN, the current network is used to evaluate the action and select $a'$, while the target network is used for evaluate the value, so that action selection is decoupled from estimation of the target:
\begin{equation}
    y \leftarrow r + \gamma Q_{\theta'} (s', a'_\ast) \qquad \textrm{ where } a'_\ast = \mathrm{argmax}_{a'} Q_{\theta} (s', a')  
\end{equation}

In practice however, it is common the case that little improvement can be gained by using double DQN, since the current and target networks are usually too similar due to slowly changed parameters in neural network models with SGD optimization. We can neither set the period of updating target too long, otherwise the derived policy would not exhibit learning and progress. As a result, double DQN does not entirely eliminate the overestimation bias. In Section \ref{sec:cq_res}, we will further experimentally show the elimination of overestimation is not effective nor sufficient in double DQN.

Instead of maintaining only two separate networks, we will use a set of $K$ models for estimating Q-values and selecting actions in our cross Q-learning. While update each network's parameters, we will calculate its TD target Q-value using one of the other $K-1$ models. More specifically, let the network with parameters we are about to adjust be our current network ($\theta_i$), and we randomly pick another network to be our target network ($\theta_j$, e.g., $j\in U[1, K]$). To compute the target Q-value, we will use the current network to evaluate the actions and select $a'$ in the next state $s'$, while the value is evaluated by using the target network, i.e., 
\begin{equation}
    y \leftarrow r + \gamma Q_{\theta_j} (s', a'_\ast) \qquad \textrm{ where } a'_\ast = \mathrm{argmax}_{a'} Q_{\theta_i} (s', a')  
\end{equation}

\begin{algorithm}[htb!]
\caption{Cross-Learning DQN}
\label{alg:onlineCrossQ}
\begin{algorithmic}[1]
\footnotesize
\STATE Initialize $K\in \mathbb{N}_+$ different Q-functions $Q(s, a|\theta^k)$ with random parameters $\theta^k$ for $k=1, \cdots, K$.
\STATE Initialize replay buffer $\mathcal{B}$.
% \REPEAT[for each episode]
\FOR{each episode \textbf{until} end of learning}
    \STATE Initialize state $s$
    % \STATE Randomly pick a network $Q^k$ to act, where $k\in \{1, \cdots, K\}$.
    % \REPEAT [for each step in an episode]
    % \For{each step in an episode}
    \FOR{step $t=1, \cdots$ \textbf{until} $s$ is terminal state of an episode} 
        % \STATE Select action $a = \textrm{argmax}_{a'} Q^k(s, a')$ with exploration
        %(with or without exploration?)
        \STATE Select action $a_t$ according to $Q$ with exploration, e.g., $a_t = \textrm{MajorityVote}\{ \textrm{argmax}_a Q_k(s, a) \}_{k=1}^K$ 
        \STATE Take action $a_t$, observe reward $r$ and next state $s'$
        \STATE Store experience tuple $<s, a_t, r, s'>$ into $\mathcal{B}$
        \STATE Sample a mini-batch of experiences from $\mathcal{B}$.
        % \STATE Randomly pick another network $Q^j$ for obtaining estimated target values.
        \FOR{all sampled experience in the mini-batch} 
            \STATE To train network $Q^i$, compute $a' = \textrm{argmax}_{a} Q^i(s', a|\theta^i)$
            \STATE Randomly pick another network $Q^j$ to estimate TD target $y = r + Q^j(s', a' | 
            \theta^j)$
            % \STATE Compute the TD error $\delta = \hat{Q} -Q^k(s, a|\theta^k)$, in which the target value is estimated using $Q^j$.
            \STATE Backpropagate TD error $\delta = y -Q^i(s, a|\theta^i)$ through $Q^i$, update $\theta^i$ with learning rate $\alpha_t$
        \ENDFOR
        \STATE $s\leftarrow s'$
    \ENDFOR
    % \UNTIL{$s$ is terminal state}
\ENDFOR
    % \UNTIL{end of learning}
\end{algorithmic}
\end{algorithm}

%We have more flexibility and options in how to utilize the $K$ copies of different Q-networks. We first express the version of direct derivation from double DQN, in which to update each network's weights, the TD target Q-value is calculated using one of the other model. 

\begin{figure}[htb]
    \centering
    \begin{subfigure}[b]{0.45\textwidth}
        \includegraphics[width=\textwidth]{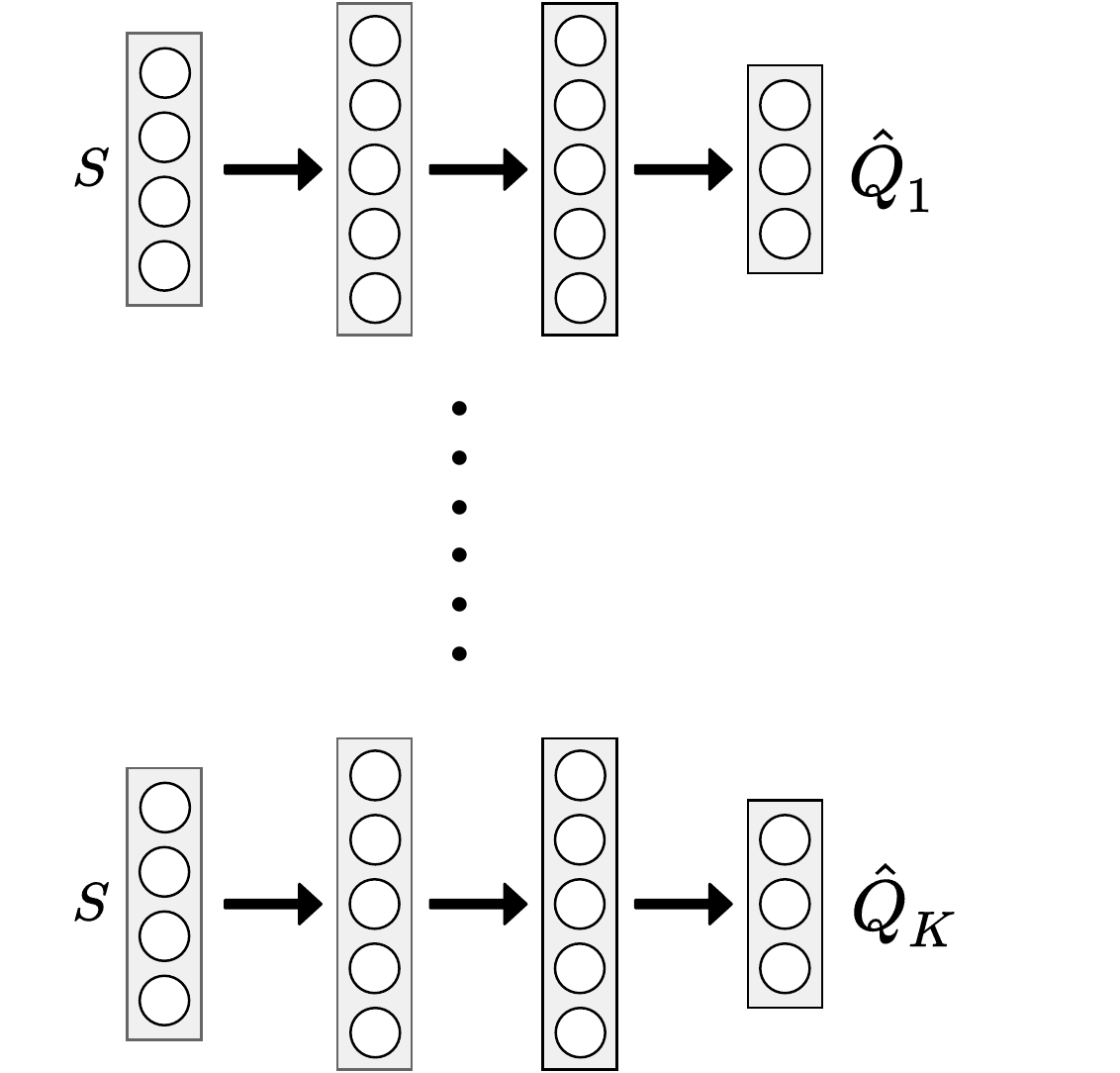}
        \label{fig:sep}
        \caption{Separated Network Design}
    \end{subfigure}
    ~
    \begin{subfigure}[b]{0.52\textwidth}
        \includegraphics[width=\textwidth]{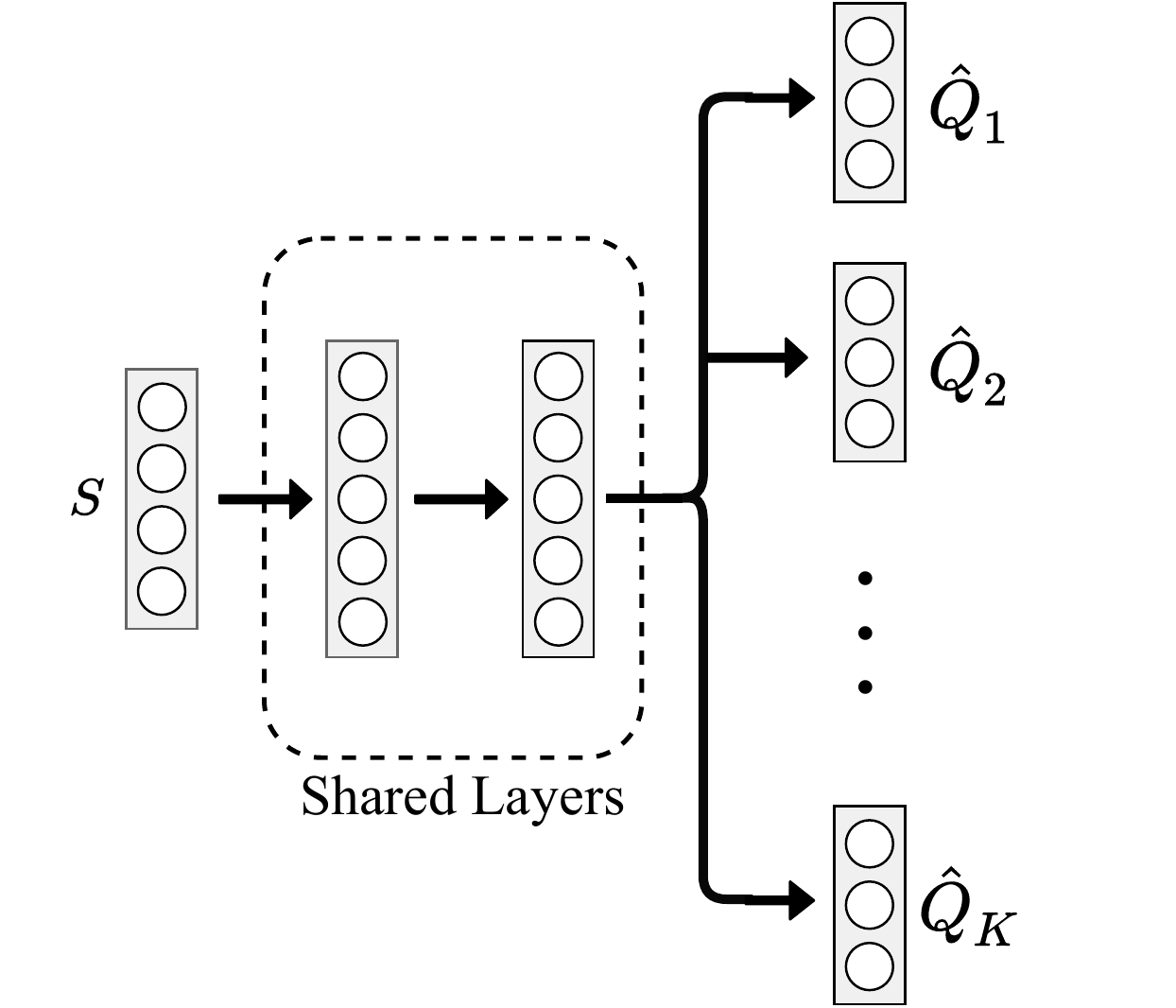}
        \label{fig:shared}
        \caption{Shared Network Design}
    \end{subfigure}
    \caption{Separate and Shared Network Architecture}
    \label{fig:arc}
\end{figure}

In implementation, we have flexibility and various options in how to utilize the $K$ different Q-networks. There always exist tradeoffs among different choices that we need to consider in order to pick the one that meets our goal most. For example, we can have different design of neural network architectures. A natural choice of having $K$ independent models is to maintain a list of separate neural networks with the same architecture, the difference between their outputs (i.e., $K$ streams of Q-values derived from the same ($s, a$)-pair as the input) comes from different random parameter initialization of each model, also is due to that different data that each model is trained upon, i.e., for each step of backpropagation, each model randomly samples a mini-batch of experiences and performs SGD optimization with the mini-batch. Maintaining $K$ copies of models implies that not only the storage for the models would be $K$ times large as a single network, also the forward propagation would take $K$ times amount of computations. 
Instead, we can utilize the shared network design for the $K$ models, in which the $K$ models shared their weights except for the last layer, which consists of $K$ value function heads from which the value functions $Q_k(s, a|\theta_k)$ are derived, and the weights on the last layer are generally different. Thus we have much less parameters in total to be trained, and the computational burden can be greatly alleviated. Moreover, as recent deep learning research reveals, the first few layers of neural network are mainly about representations learning, the shared layers provide the same features expressed for computing $Q$, this can be seen as online transfer of learned knowledge among models. Note that in shared learning settings, in order to avoid premature learning and suboptimal convergence, the gradients of the network except the last layer are usually normalized by $1/K$, but this also results in slower learning early on. 
On the other hand, the separate models are simpler yet provide more variability in $Q$-values, also are more stable during training. In addition, when we train the networks in distributed system, the separate networks do not depend on others' weights thus can be learned independently, which requires much less information exchange and this could be a huge advantage for distributed learning. The comparison of the separate and shared network architectural design is shown in Figure \ref{fig:arc}.

% Another choice we can make is that how to combine action selections into a single policy w
With $K$ different models (or heads), while each could derive a possibly different policy, there is no doubt that during test phase we should take advantage of ensembles, for instance by choosing the action with the majority votes across the outputs. However, we can make choices on how to combine action selections into a single policy during training. With ensemble action selection such as majority voting, the derived policy is often superior than any individual one, thus greatly reduces the variance during training, as we will experimentally show in Section \ref{sec:cq_res}. This in turn refines exploitation, results in great variance reduction of $Q$-values and speeds up learning. Note that to deal with exploration-exploitation dilemma, $\varepsilon$-greedy strategy is needed to encourage exploration. 
On the other hand, we may also randomly pick a single network from the $K$ models, and act as it suggests during training. This falls into the paradigm of Bootstrapped DQN \cite{osband2016bdqn}, which encourages exploration, in the cost of slower early learning (see Section \ref{sec:cq_res}), but may learn better policy later with more exploration. Another advantage of bootstrapped action selection is that it can slightly reduce computational burden, since instead of forward passing and computing all $K$ of the $Q$-values for action selection, we can calculate only one of them. The procedure of bootstrapped version of cross DQN is presented in Algorithm \ref{alg:bcdqn}.

\begin{algorithm}[htb!]
\caption{Bootstrapped Cross DQN}
\label{alg:bcdqn}
\begin{algorithmic}[1]
\footnotesize
\STATE Initialize $K\in \mathbb{N}_+$ different Q-functions $Q(s, a|\theta^k)$ with random parameters $\theta^k$ for $k=1, \cdots, K$.
\STATE Initialize replay buffer $\mathcal{B}$.
% \REPEAT[for each episode]
\FOR{each episode \textbf{until} end of learning}
    \STATE Initialize state $s$
    \STATE Randomly pick a network $Q^k$ to act, where $k\in \{1, \cdots, K\}$.
    % \REPEAT [for each step in an episode]
    % \For{each step in an episode}
    \FOR{step $t=1, \cdots$ \textbf{until} $s$ is terminal state of an episode} 
        \STATE Select action $a_t = \textrm{argmax}_{a'} Q^k(s, a')$ with exploration
        % \STATE Select action  $a = \textrm{MajorityVote}\{ \textrm{argmax}_a Q_k(s, a) \}_{k=1}^K$  with exploration 
        \STATE Take action $a_t$, observe reward $r$ and next state $s'$
        \STATE Store experience tuple $<s, a_t, r, s'>$ into $\mathcal{B}$
        \STATE Sample a mini-batch of experiences from $\mathcal{B}$.
        % \STATE Randomly pick another network $Q^j$ for obtaining estimated target values.
        \FOR{all sampled experience in the mini-batch} 
            \STATE To train network $Q^i$, compute $a' = \textrm{argmax}_{a'} Q^i(s', a'|\theta^k)$
            \STATE Randomly pick another network $Q^j$ to stimate TD target $y = r + Q^j(s', a' |
            \theta^j)$
            % \STATE Compute the TD error $\delta = \hat{Q} -Q^k(s, a|\theta^k)$, in which the target value is estimated using $Q^j$.
            \STATE Backpropagate TD error $\delta = y -Q^i(s, a_t|\theta^i)$ through $Q^i$, update $\theta^i$ with learning rate $\alpha_t$
        \ENDFOR
        \STATE $s\leftarrow s'$
    \ENDFOR
    % \UNTIL{$s$ is terminal state}
\ENDFOR
    % \UNTIL{end of learning}
\end{algorithmic}
\end{algorithm}

Another choice we can make is the training frequency. %, which usually also serves as a hyperparameter in DQN training. 
In our cross DQN settings, when backpropagation occurs, we can either choose to train on a single network (e.g., the single model that provides the action selection), or each of the $K$ networks could independently sample a mini-batch of experiences and perform SGD optimization. The latter would increase the sample efficiency and speed up learning, while the former would reduce the computational burden, in which the number of backpropagation (which is the most computational expensive) remains the same as in a single DQN. In addition, with the former setting, 
our cross Q-learning does not require maintaining copies of the networks as the target. Experimentally, we found that freezing targets merely has any effect on stabilization of learning, but only costs doubled memory for model storage. This is due to two reasons. First, 
we bootstrap a model that is different than the current one, when $K\ge 2$, the variety of models ensures the difference in parameter initialization, as well as the difference of mini-batch data their learning based upon, which in turn ensures the independence of Q-value estimates. Secondly, with less frequent update of each network, the bootstrapped target $Q$-value changes less as well, also helps stabilize learning.

\section{Experimental Results}
\label{sec:cq_res}

In this paper, we conducted experiments on two classical control problems, CartPole and LunarLander, for extended tests. We selected these testbeds in the aim of covering different challenges, especially in terms of complexity. As both environments interfaced through OpenAI gym environment \cite{gym}, unless specified otherwise. 
The neural networks have a number of hyperparameters. The combinatorial space of hyperparameters is too large for an exhaustive search, therefore we have performed limited tuning. For each component, we started with the same settings as in \cite{baselines} in order to make comparisons with states of the art results. 
 
\subsection{CartPole}
\subsubsection{Experimental Setup}
The CartPole, also known as an inverted pendulum, in which a pole (or pendulum) is attached by an un-actuated joint to a cart (i.e., the pivot point). The pendulum starts upright at the center of a 2D track but is unstable since the center of gravity is above the pivot point. The goal of this task is to keep the pole balanced and prevent it from falling over, by applying appropriate force to the pivot point, while the force could move the cart along the frictionless track with finite length of 4.8 units. An immediate reward of $+1$ is provided for every timestep that the pole remains not falling over, and the maximum cumulative rewards in an episode are clipped to 200. An episode also ends when the pole is slanted with degree $> 15^\circ$ from vertical, or the cart moves out of the track \cite{barto1983neuronlike}. In each timestep, the agent is provided with current state $s\in \mathbb{R}^4$, which represents cart position, cart velocity, pole angle, and pole angular velocity, respectively. A unit force either from left or right can be applied, thus the actions are discrete with $a\in \{-1, 0, +1\}$. 

As in \cite{baselines}, we approximate the $Q$-values using a neural network with two fully-connected hidden layers (which consist of 64 and 32 neurons, respectively). 
We train each of the neural networks for 1000 episodes (approximately a little less than 200000 steps),
with a FIFO memory of size $5\times 10^4$ transitions for experience replay. A target network is updated every 500 steps to further stabilize learning. The adaptive moment estimation (Adam) optimizer with learning rate $\alpha=0.001$ is used to train the network, since it is in general less sensitive to the choice of the learning rate than other stochastic gradient descent algorithms \cite{kingma2014adam}. 
The optimization is performed on mini-batches of size 32, sampled uniformly from the experience replay. 
The discount factor $\gamma$ is set to 0.99, and $\varepsilon$-greedy policy is used for choosing actions throughout interacting with the environment, which starts with exploration $\varepsilon=1$, and annealed to $0.02$ in the first 10000 steps. 

After every 20 training episodes, we conduct a performance test that plays 10 full episodes using the greedy policy deterministically derived from the current network. For the models with $K > 1$, majority voting is used for the action selection disregard whether or not bootstrapped $Q$-value head is used during training. The cumulative rewards of each test episode are used for comparison among different models. 
Moreover, in order to comparing the estimation of $Q$-values among models, every 20 training episodes, we randomly sample a batch of historical $1024$ $(s, a)$-pairs from the replay buffer and compute their $Q$-values using current network. More than one thousand samples ensure that their mean is somewhat representative for $Q$-values under current model.

\subsubsection{Analysis of Cross Q-learning Effects}
\label{sec:cp12510}

\begin{figure*}[!htb]
    \centering
    \begin{subfigure}[b]{\textwidth}
        \includegraphics[width=\textwidth]{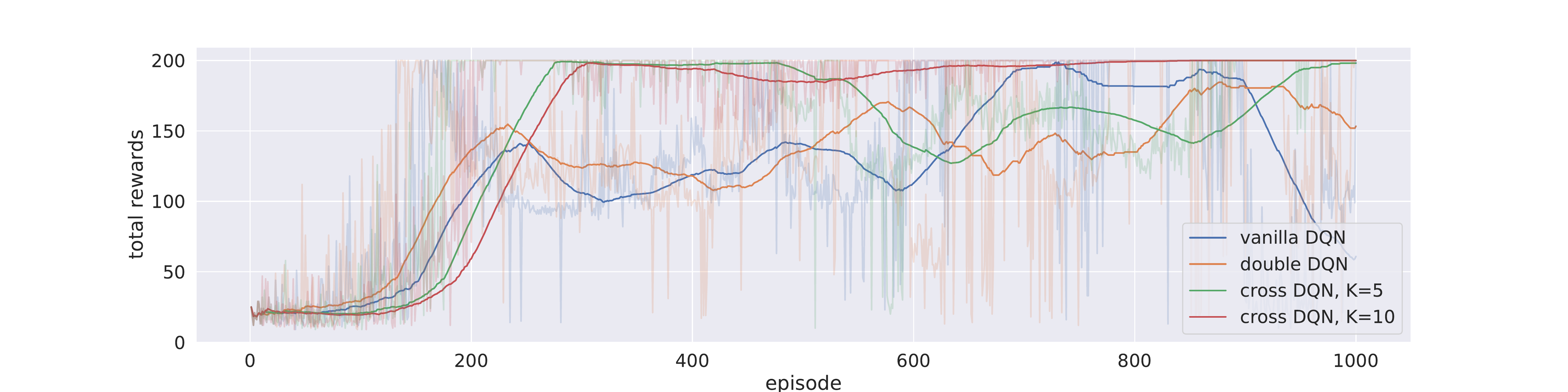}
        \caption{Learning curves}
    \end{subfigure}
    ~
    \vspace{.2in}
    \begin{subfigure}[b]{\textwidth}
        \includegraphics[width=\textwidth]{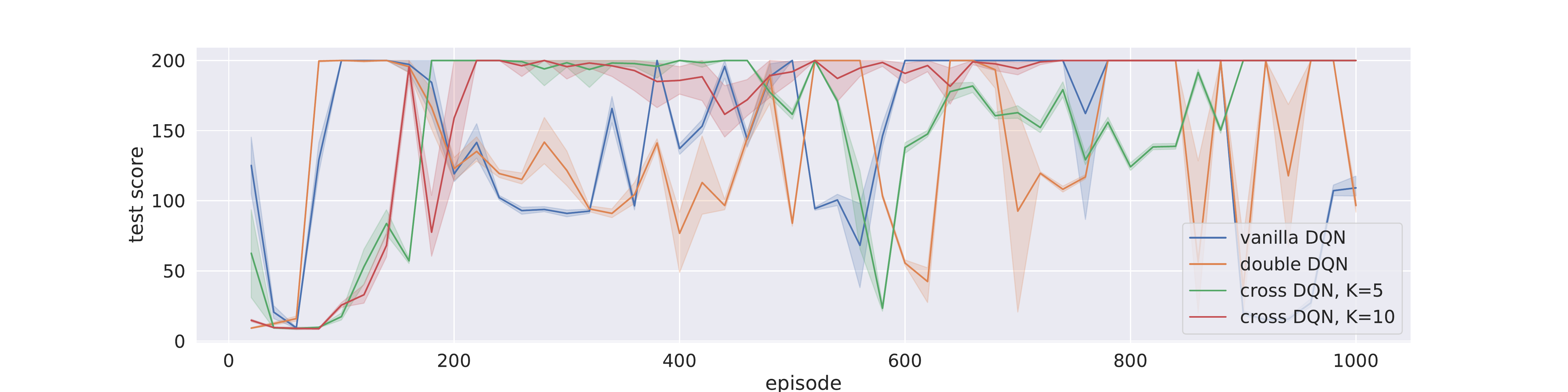}
        \caption{Model test performance. Every 20 training episodes, 10 full test episodes were conducted.}
    \end{subfigure}
    ~
    \vspace{.2in}
    \begin{subfigure}[b]{\textwidth}
        \includegraphics[width=\textwidth]{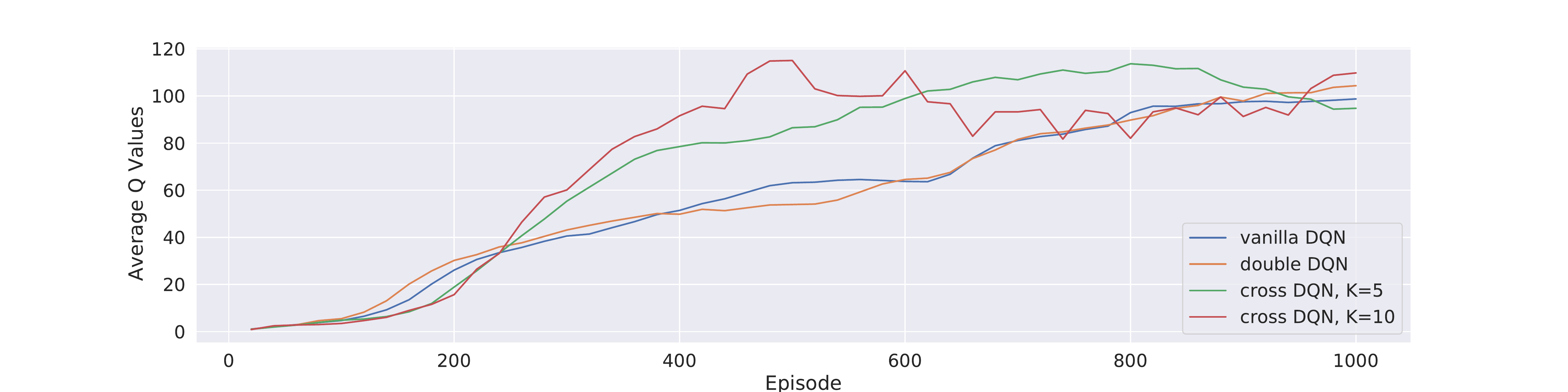}
        \vspace{-.3in}
        \caption{Mean of Q-value estimations on \textit{CartPole}. Every 20 training episodes, 1024 $(s, a)$-pairs were bootstrapped.}
    \end{subfigure}
    \caption{Comparison of vanillar DQN, double DQN and 
    cross DQNs of $K=5$, $K=10$ on \textit{CartPole}.  }
    \label{fig:cp12510}
\end{figure*}

We compared our cross Q-learning algorithms with vanilla DQN and double DQN. Note that vanilla DQN uses single estimators, while double DQN uses double estimators, and our cross DQN uses cross estimators. $K=5$ and $K=10$ are used in cross DQNs. Figure \ref{fig:cp12510}(a) illustrate the training history of episodic total rewards of the four models, from which we can see that although with a single network (vanilla and double DQNs), the agent starts to learn early on with less samples, in particular, double Q-learning helps the single network to learn even faster, however, the learned models are not stable. With cross Q-learning, although the networks learn slower at the beginning, in particular, cross DQN with $K=10$ started to learn even later than cross DQN with $K=5$, once cross DQNs start to learn, the performance improvement is substantial. Not only the total rewards are higher, the learning is also much more stable. After 300 episodes, the training total rewards converge to 200 for $K=10$ cross DQN, with little variation (due to $\varepsilon$ exploration). $K=5$ cross DQN has more variation, but it also seems to converge after 900 episodes, while vanilla DQN and double DQN are easily deteriorated, and have much larger variations.

% \begin{figure*}[h!]
%     \centering
%     \begin{subfigure}[b]{\textwidth}
%         \includegraphics[width=\textwidth]{crossQ/figs/cartpole/train12510.png}
%         \vspace{-.2in}
%         \caption{Total rewards of each episode}
%     \end{subfigure}
%     ~
%     \begin{subfigure}[b]{\textwidth}
%         \includegraphics[width=\textwidth]{crossQ/figs/cartpole/train_smooth12510.png}
%         \vspace{-.2in}
%  %       \caption{}
%         \caption{smoothed version}
%     \end{subfigure}
%     \caption{Learning curves of vanilla DQN (green), double DQN (purple) v.s. Cross DQN of $K=5$ (yellow), $K=10$ (brown) on \textit{CartPole}}
%     \label{fig:cp12510train}
% \end{figure*}

The performance improvement can be more clearly seen in Figure \ref{fig:cp12510}(b). After 300 episodes of training, the policies derived cross DQN with $K=10$ become more and more stable, the variance of test total rewards become zero close to the end of training. Cross DQN with $K=5$ deteriorates after 500 episodes of training, but later it also learns to derive stable policy that has total rewards of 200 with tiny variances. Whereas the policies derived from vanilla DQN and double DQN can only get score which is approximately half of cross DQNs, and with large variances. The policy derived from double DQN seems to be a little better than that from vanilla DQN, but the improvement is not as significant as that of using cross Q-learning. 

Furthermore, part of the reason for slower start of cross DQN is due to our learning settings, in which we only perform SGD optimization on one of the networks (or heads). In other words, we reduce the learning frequency of each network (or head) down to $1/K$ to alleviate the computational effort, at the cost of slower start on learning. If we increase the learning frequency (i.e., backpropagate for each of the $K$ networks/heads every time), the learning should be faster. 

We also plot the average $Q$-values from bootstrapped 1024 $(s, a)$-pairs as shown in Figure \ref{fig:cp12510}(c). We observe that the beginning of learning, vanilla DQN has highest estimates of $Q$-values, which is an evidence of overestimation. The estimates from double DQN is lower, but only for limited amount, therefore we say that double Q-learning may have not solve the overestimation problem completely. Cross DQNs have quite smaller estimations at the beginning, in particular, as $K$ gets larger, the estimates of $Q$-values become even lower. Overestimation is clearly an obstacle of effective learning, as a result, the estimated $Q$-values from cross DQNs are substantially higher than that from vanillar or double DQNs, since cross DQNs has derived better policies and obtained higher rewards. The $Q$-values estimates from cross DQNs start to converge after the derived policies stabilized, 
At the end of training, the estimated $Q$-values from the four different models are about at the same level, however, note that the estimates from vanilla and double DQNs continue increasing, and their derived policies are not stable, also have lower rewards. Our cross Q-learning algorithm has addressed the overestimation problem better. 

\subsubsection{Effects of dueling DQN \& Bootstrapped DQN}

\begin{figure*}[htb]
    \centering
    \begin{subfigure}[b]{\textwidth}
        \includegraphics[width=\textwidth]{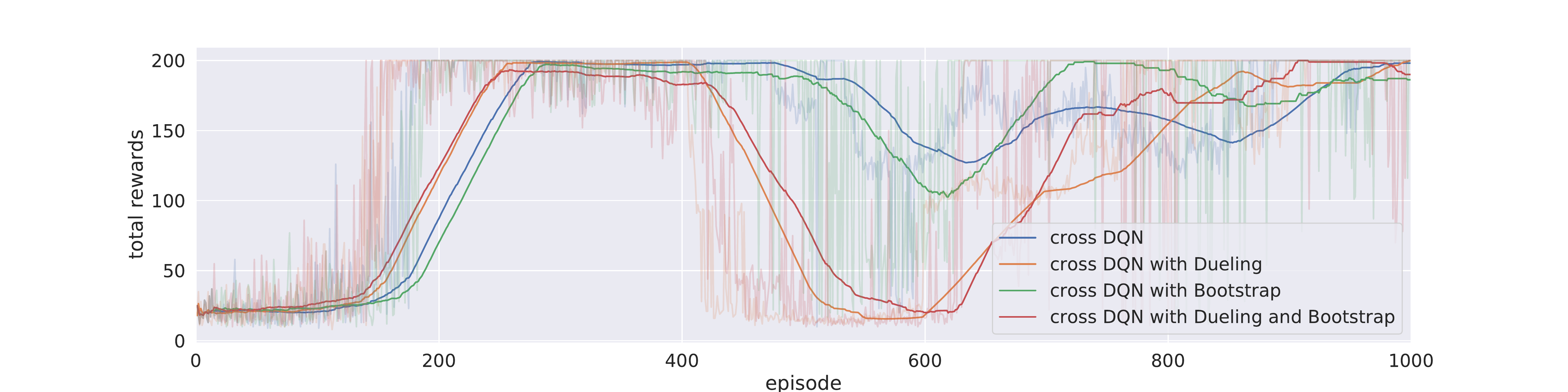}
        % \vspace{-.2in}
        \caption{Learning curves}
    \end{subfigure}
    ~
    \begin{subfigure}[b]{\textwidth}
        \includegraphics[width=\textwidth]{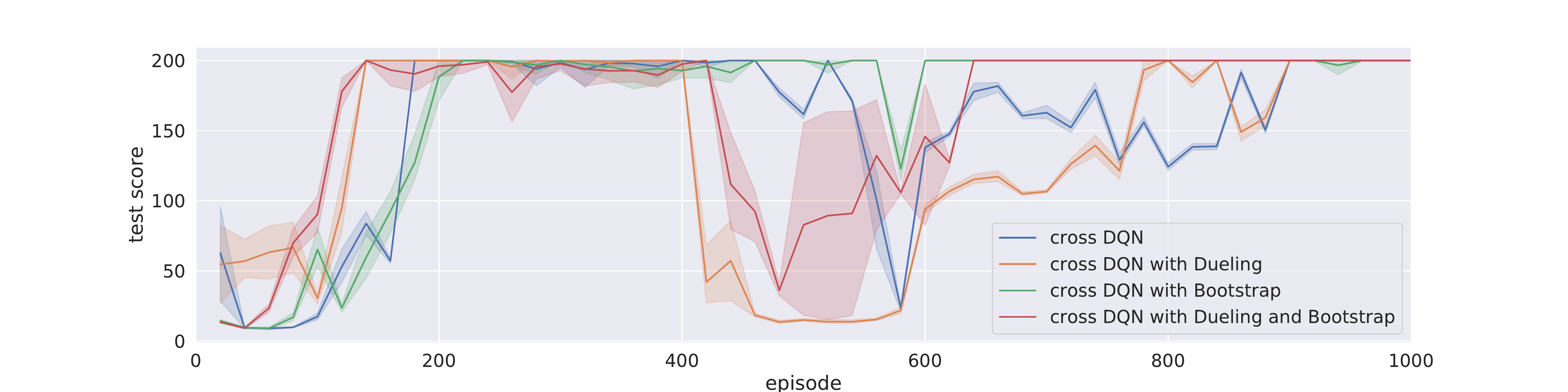}
        % \vspace{-.2in}
 %       \caption{}
    \caption{Model test performance. Every 20 training episodes, 10 full test episodes were conducted.}
    \end{subfigure}
    \caption{Comparison of cross DQNs of $K=5$. Cross DQN with ensemble voting, with dueling DQN and voting, with bootstrapped DQN, and with both dueling \& bootstrapped DQN on \textit{CartPole}.  }
    \label{fig:cp5}
\end{figure*}

\begin{figure*}[htb]
    \centering
    \begin{subfigure}[b]{\textwidth}
        \includegraphics[width=\textwidth]{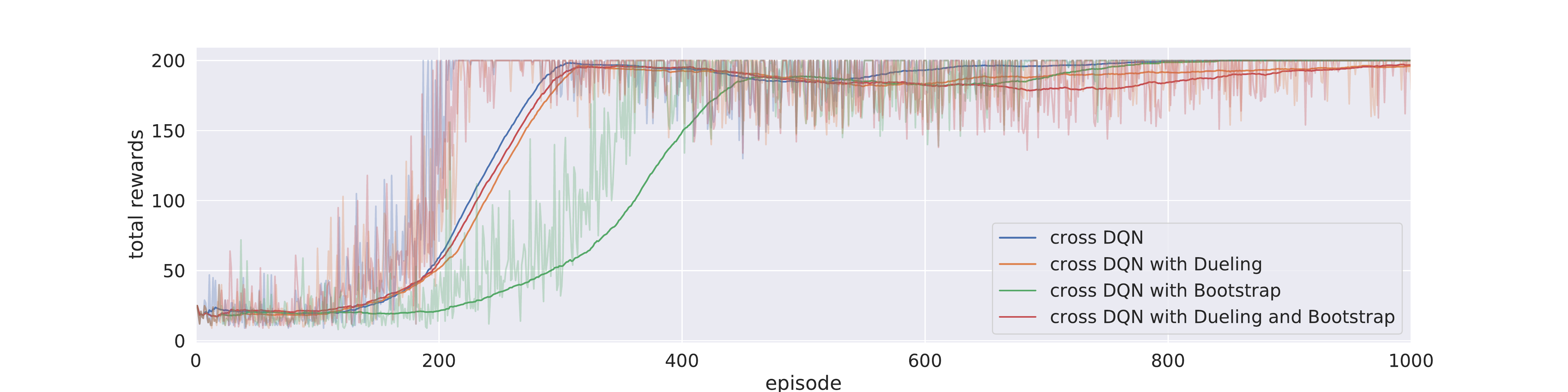}
        % \vspace{-.2in}
        \caption{Learning curves}
    \end{subfigure}
    ~
    \begin{subfigure}[b]{\textwidth}
        \includegraphics[width=\textwidth]{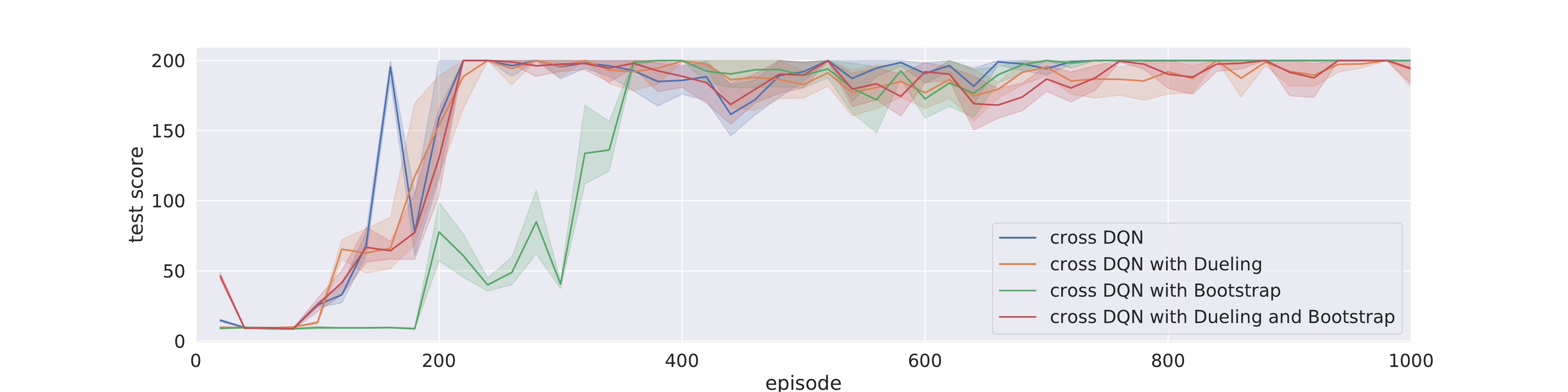}
        % \vspace{-.2in}
        \caption{Model test performance. Every 20 training episodes, 10 full test episodes were conducted.}
    \end{subfigure}
    \caption{Comparison of cross DQNs of $K=10$. Cross DQN, with dueling DQN, with bootstrapped DQN, with both dueling \& bootstrapped DQN on \textit{CartPole}.  }
    \label{fig:cp10}
\end{figure*}

As the cross learning architecture shares the same input-output interface with standard DQN, we can recycle many recent advances in DQN research.
We have mentioned one variant in Section \ref{sec:cq_cross} that it can combined with Bootstrapped DQN for action selection during training, while in Secction \ref{sec:cp12510}, our experiments for cross DQN are based on majority voting from $K$ different $Q$-functions. Furthermore, it is convenient to combine the dueling architecture into each of the $K$ networks. The goal of dueling DQN is to reduce variance for $Q$-value estimation, by subtracting a baseline and emphasizing the advantages among different actions, thus accelerates learning effectively. The variance reduction is performed on a single network's estimation, while our cross Q-learning reduces variance from a different perspective. For each network, the target values were calculated with other models by bootstrapping from multiple $Q$-values, thus introduces some bias. Due to the bias-variance tradeoff, however, the variance of our estimates decreases, and thus the overall error becomes smaller. In addition, the maximum operator induces overestimation bias, while cross-estimator tends to introduce bias in the other direction, thus greatly alleviates overestimation problem. 

Figure \ref{fig:cp5} and Figure \ref{fig:cp10} illustrate the training and testing performance of cross DQN with different architectures, for the cases of $K=5$ and $K=10$, respectively. We can see that dueling architecture speeds up early on learning effectively, without hurting the model performance later in general. On the other hand, Bootstrapped DQN slows learning at beginning, especially when $K$ is large, since the selected actions varies among networks at beginning quite a bit.
For example, the $K=10$ cross DQN with bootstrap converges around 400 episodes while the other cross learning agents converges before 200 episodes. 
But after learned something, the bootstrapped action selection won't hurt the model. In fact, it might help learning for more complicated tasks because of more exploration early on. 
At least, using bootstrapped DQN can help our cross DQN agent make faster action selection during training and reduce computational burden slightly, since instead of calculate all $K$ Q-values, we can calculate only one of them. Moreover, by comparing the learning curves of bootstrapped cross DQNs with different $K$s, we can conclude that it is primarily our cross Q-learning rather than policy ensemble that greatly reduces the variance, as with $K=10$ the variations are much smaller that that with $K=5$, though policy ensemble further reduces the variance greatly, and during testing phase, our agent can definitely benefit from ensemble of multiple models. Naturally combined crossed Q-learning with dueling and bootstrapped DQN, our model aggregates the merits from all three perspectives.

\subsection{Lunar Lander}

The task of Lunar Lander in Box2D \cite{catto2011box2d} is to land the spaceship between the flags smoothly. 
In each step, the agent is provided with the current state $s$ of the lander in $\mathbb{R}^8$, in which 6 of the dimensions are in continuous space whereas the other 2 are dummy variables in discrete space, and the agent is allowed to make one of the 4 possible actions (i.e., the action space is discrete): fire the left, right, or down throttle so that the lander could obtain a force toward the opposite direction, or do nothing. At the end of each step, the agent receives a reward and moves to a new state $s'$. An episode finishes if the lander rest on the ground at zero speed (receives additional reward of $+100$), or hits the ground and crashes (receives additional $-100$ reward), or flies outside the screen, or reaches the maximum of 1000 time steps of one episode. The agent aims for successful landing which is defined as reaching the landing pad (between two flags) centered at the ground at the speed of zero, and receives an additional reward in range  $[100, 140]$, while landing outside the pad would cause some penalty.

We built each network with two fully-connected hidden layers, which consist of 128 and 64 neurons, respectively. 
We train each of the neural networks for 10000 episodes for the LunarLander task, with a much larger replay buffer of size $10^6$. The target network update is set to every 1000 steps for vanilla and double DQN, and learning rate $\alpha=0.001$ and batch size of 64 are used for Adam optimizer to train all the models. 
The discount factor $\gamma$ is again 0.99, and 
exploration rate $\varepsilon$ is set to annealed to $0.02$ in the first 100000 steps. 
And again, $Q$-values for bootstrapped 1024 $(s, a)$-pairs are evaluated and 
10 episodes of performance tests with current policy are conducted every 20 training episodes.

\begin{figure*}[!htb]
    \centering
    \begin{subfigure}[b]{\textwidth}
        \includegraphics[width=\textwidth]{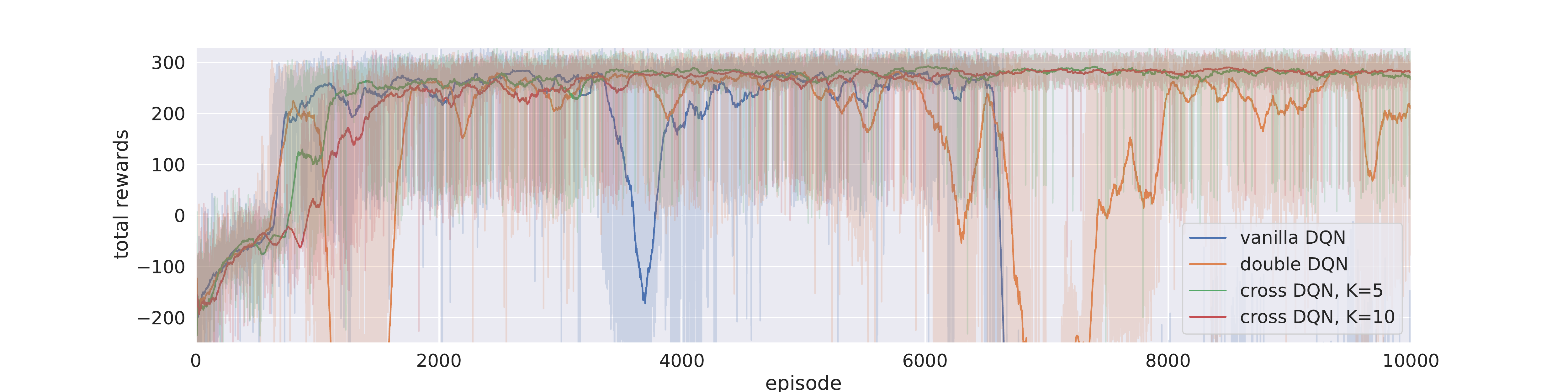}
        % \vspace{-.3in}
        \caption{Learning curves}% (will redraw as exponentiall moving average smoothing instead of simple MA smoothing)}
    \end{subfigure}
    ~
    \vspace{.2in}
    \begin{subfigure}[b]{\textwidth}
        \includegraphics[width=\textwidth]{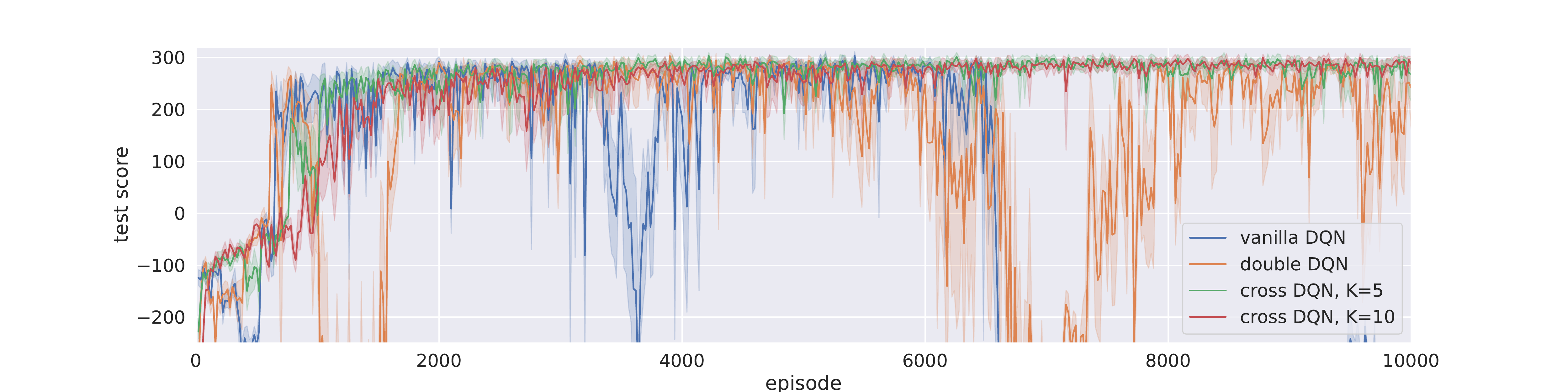}
        % \vspace{-.3in}
        \caption{Model test performance. Every 20 training episodes, 10 full test episodes were conducted.}
    \end{subfigure}
    ~
    \vspace{.2in}
    \begin{subfigure}[b]{\textwidth}
        \includegraphics[width=\textwidth]{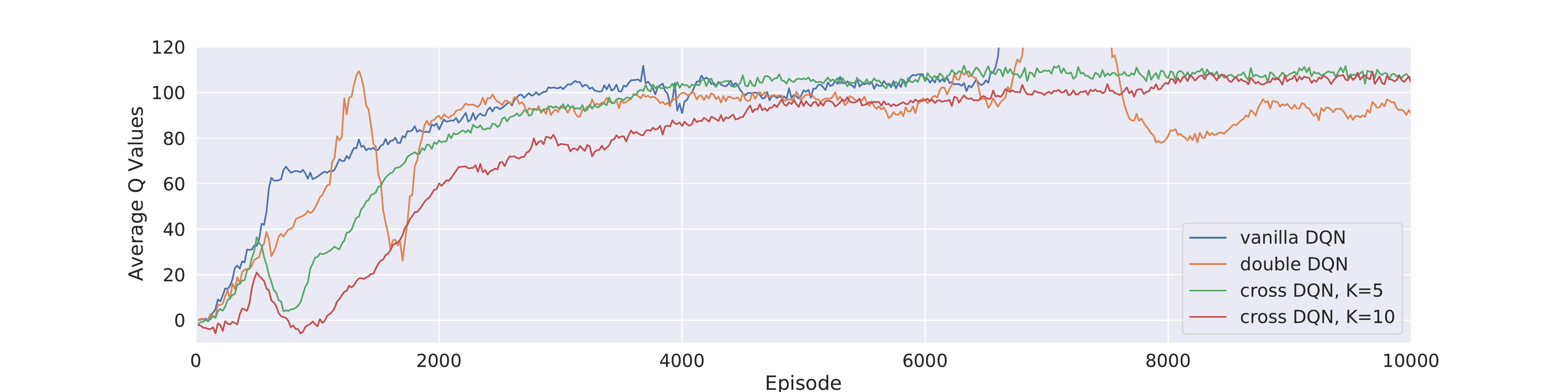}
        % \vspace{-.3in}
        \caption{Mean of Q-value estimations on LunarLander. Every 20 training episodes, 1024 $(s, a)$-pairs were bootstrapped.}
    \end{subfigure}
    \caption{Comparison of vanillar DQN, double DQN and 
    cross DQNs of $K=5$, $K=10$ on \textit{LunarLander}.  }
    \label{fig:lander12510}
\end{figure*}

In Figure \ref{fig:lander12510}, We compared our cross Q-learning algorithms with vanilla DQN and double DQN. 
With slower learning in the first a few hundreds of episodes due to our experimental design of the learning frequencies, cross DQNs learned much better and more stable policies, while vanilla and double DQN have large variances in both learning curves and performance testing. 
Figure \ref{fig:lander12510}(c) clearly shows that from the beginning, vanilla DQN optimistically gathers the occasional large rewards which are due to the high variance, and produces great overestimations. Double DQN slightly allivates the problem, but cannot avoid the overestimation effectively. The derived policies from these two networks are then not optimal nor stable. As learning going on, the estimated $Q$-values from both vanilla and double DQN explode, resulting in that the derived policies are no better than random actions. On the other hand, cross DQNs have much lower $Q$-value estimations at the beginning, and the estimates from model with $K=10$ are even lower than that from model with $K=5$.   

After %getting rid of a local optima (hovering), 
1000 episodes, the estimates continue growing until convergence, and their values converge to a same level at about $105$. The derived policies are very stable, with total rewards close to 300 and also have little variance. Note that double DQN has lower estimates of $Q$-values than cross DQNs after 8000 episodes of training. The reason is that the corresponding policies from double DQN are much worse, and it does not indicate that double DQN addresses overestimation better.

\begin{figure*}[htb]
    \centering
    \begin{subfigure}[b]{\textwidth}
        \includegraphics[width=\textwidth]{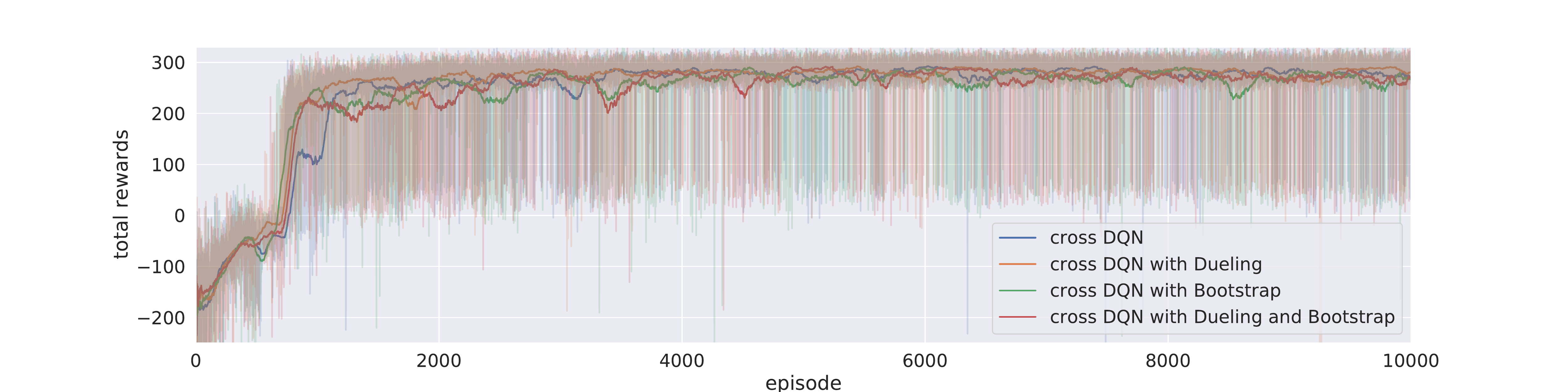}
        % \vspace{-.2in}
        \caption{Learning curves}
    \end{subfigure}
    ~
    \begin{subfigure}[b]{\textwidth}
        \includegraphics[width=\textwidth]{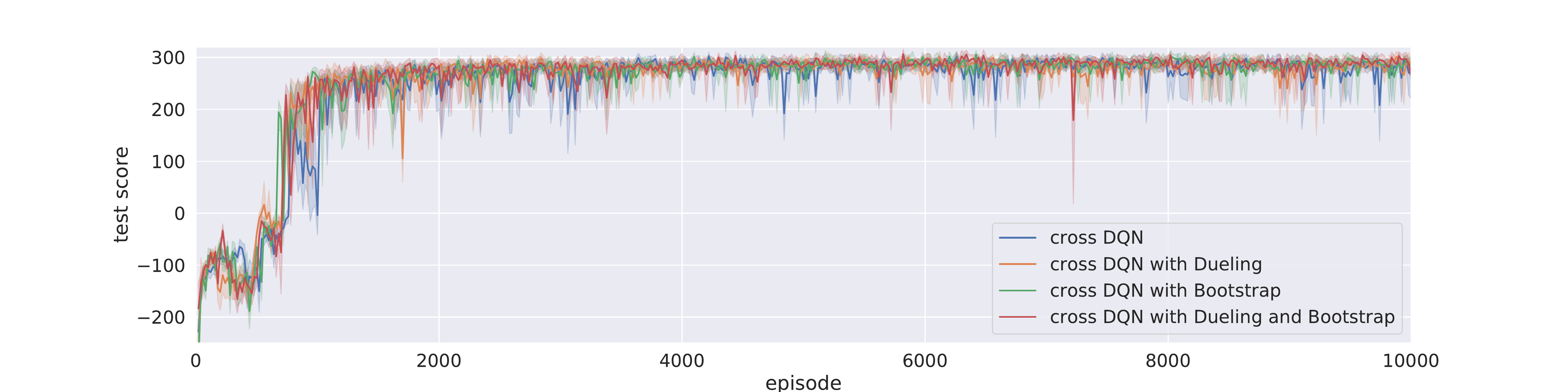}
        % \vspace{-.2in}
        \caption{Model test performance. Every 20 training episodes, 10 full test episodes were conducted.}
    \end{subfigure}
    \caption{Comparison of cross DQNs of $K=5$. Cross DQN, with dueling DQN, with bootstrapped DQN, and with both dueling \& bootstrapped DQN  on \textit{LunarLander}.  }
    \label{fig:lander5}
\end{figure*}

\begin{figure*}[htb!]
    \centering
    \begin{subfigure}[b]{\textwidth}
        \includegraphics[width=\textwidth]{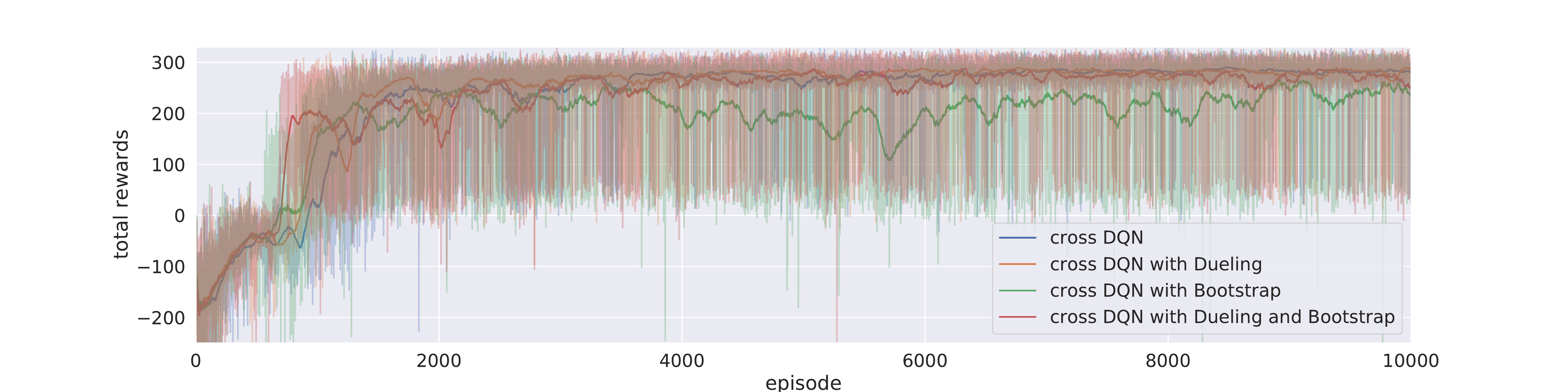}
        % \vspace{-.2in}
        \caption{Learning curves}
    \end{subfigure}
    ~
    \begin{subfigure}[b]{\textwidth}
        \includegraphics[width=\textwidth]{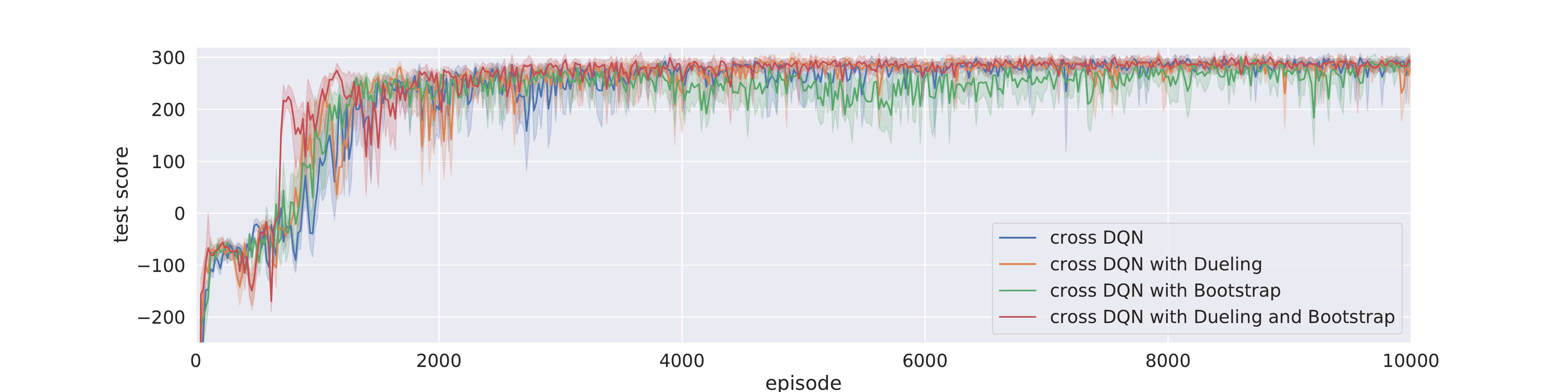}
        % \vspace{-.2in}
        \caption{Model test performance. Every 20 training episodes, 10 full test episodes were conducted.}
    \end{subfigure}
    \caption{Comparison of cross DQNs of $K=10$. Cross DQN, with dueling DQN, with bootstrapped DQN, and with both dueling \& bootstrapped DQN  on \textit{LunarLander}.  }
    \label{fig:lander10}
\end{figure*}

Comparing Figure \ref{fig:lander5} and Figure \ref{fig:lander10}, $K=5$ seems works even better than $K=10$ for most of time. Especially for $K=10$ bootstrapped cross DQN, both the learning curve and the test scores are lower than other cross DQN models. This indicates that it is not always the larger $K$ the better, since cross estimator would induce underestimate bias, and too much underestimation may also hide the real better actions and thus hurt the model performance. In fact, $K=10$ cross DQN might have too much underestimation at the beginning, which slows down the learning process significantly. 
But overall, the $K=10$ bootstrapped cross learning with dueling architecture performs best among all models, including all $K=5$ cross DQNs. We say that the DQN architectures are too complicated, and the aggregated effect may significantly change the performance of a particular model. Generally speaking, our cross DQNs favor underestimation, which should be much better than overestimation if no unbiased estimation can be achieved, since underestimations do not tend to propagate too much during training, as lower valued actions are avoided by the greedy action selection mechanism. And the bias-variance tradeoff tells us that the overall error can be reduced when the variance of our estimates is greatly decreased, by introducing slight negative bias, this in tern leads to better model performance.

Note that the derived policies from cross DQNs are much more stable in general, and hard to deteriorate. There are at least two reasons for this phenomena. First, cross Q-learning effectively addressed overestimation problem, thus premature policy would be more difficult to derived from cross DQN. In addition, we always ensemble policies using methods such as majority voting during test time, which in general is superior and has a stabilizing effect for action selections. The improved stability comes from larger barrier for altering the decision boundaries, and we could care much less about the early termination as an additional hyperparameter during training. This is yet another advantage of using multiple networks as in cross DQN.

% \FloatBarrier
\section{Conclusions and Future Work}
\label{sec:cq_conc}
In this paper, we have presented the cross Q-learning algorithm, an extension to DQN that effectively reduces overestimation, stabilizes training, and improves performance. 
Cross DQN is a simple extension that can be easily integrated with other algorithmic improvement such as dueling network and bootstrapped DQN, leads to dramatic performance enhancement.
We have both shown in theory and demonstrated in several experiments of classical control problems that the proposed scheme is superior in reducing overestimation and leads to better policies derivation, compared to widely used approaches such as double DQN. 
Cross learning favors underestimation, the introduced negative bias can greatly help variance reduction. We analyze this effect from the famous bias-variance tradeoff point of view. However, this also indicates that it is not the case the larger $K$ the better model performance in cross DQN. Nevertheless, DQN models tolerate underestimation much more than overestimation, as lower valued actions can be avoided by the greedy action selection mechanism. 

It is noted that the computation complexity of cross DQN is generally higher, comparing with that of single network DQNs. We can, however, greatly reduce the complexity given the flexibility provided by our model.
In addition, ensemble policies from multiple networks help stabilize the decision space, which can be utilized optionally in stablizing learning and definitely during testing. 

As future work, we would apply cross learning to the state-of-the-art actor-critic methods in continuous control, further reduce the overestimation and stabilize those algorithms. Also, analysis from statistical learning theory could be helpful for us to derive more advanced cross learning strategies, for instance, better bootstrap estimations may be obtained by mimicking the $K$-fold cross validation \cite{van2013estimating}, or from Bayesian perspective \cite{d2016estimating}. 

Moreover, it worth noting that in each step of Q-learning (and more general value-based RL), we utilize $Q$-values in several different places. Now that a set of $K$ different $Q$-functions are applied, we can make different choices for picking particular one to use. We call them generalized cross learning in DQNs, and some existing work can be fell into a particular subclass of our generalized method. 
The first place that $Q$-values are utilized is 
when the agent makes decision for choosing an action $a_t$ at time step $t$ while observing $s_t$. 
We can pick a random $Q$-function for action selection, and this is exactly what bootstrapped DQN \cite{osband2016bdqn} does. We say the bootstrapped DQN is a special case of our generalized cross DQN.
The next place is at TD update when the target Q-values need to be evaluated for choosing the next action $a'$, which might not be executed, but is used to evaluate the current target Q-value and derive the $\max$ operator. Recall in Q-learning we use the maximum estimator.  
Finally, after picking the next action $a'$, its value can be evaluated, again we have choices here for picking a $Q$-function to use. 
In the version of our cross DQN we presented in this work, which is directly derived from double DQN, 
we decoupled the selection and evaluation of the next action $a'$, where the current network is used for evaluating $a'$ while another target network is used for selecting $a'$.
We could try to do the opposite in certain circumstances, i.e., select $a'$ with the current network and bootstrap another network to evaluate $a'$, which should have the effect of decrease bias but increase variance due to bias-variance tradeoff in general statistical learning scheme.
One can further analyze and experiment with other generalized cross Q-learning variants. 

\bibliographystyle{unsrt}
\bibliography{main.bbl}

\newpage
\appendix

% \begin{appendices}
\section{Estimating the Maximum Expected Values}
\label{sec:cq_est}
For Q-learning, the action is selected according to the estimated target Q-values. This is an instance of a more general maximum expected value estimation problem, which is formed as follows.
Consider a set of $|\mathcal{A}|$ random variables $Q = \{Q_{a_1}, \cdots, Q_{a_{|\mathcal{A}|}} \}$, we are interested in finding the maximum expected value among the set of variables, which is defined as $$\max_a \mu_a = \max_a \mathbb{E} [Q_a]$$
while each $\mathbb{E} [Q_a]$ is usually estimated from samples. Let $\Omega_a$ denote the sample space for estimating $Q_a$, for $a\in \mathcal{A}$, and we further assume that the samples in $\Omega_a$ are i.i.d.
The sample mean $\hat{\mu}_a = \frac{1}{|\Omega_a|} \sum_{x\in \Omega_a} x$ is then an unbiased estimator for $\mathbb{E} [Q_a]$.

Let $f_a: \mathbb{R} \to \mathbb{R}$ be the probability density function (PDF) for the variable $Q_a$, and $F_a (x) = \int_{-\infty}^x f_a(x)dx$ be the cumulative density function (CDF). The maximum expected value is then 
\begin{equation}
    \label{eqn:maxe}
    \max_a \mathbb{E} [Q_a] = \max_a \int_{-\infty}^\infty x f_a(x) dx .
\end{equation}

\subsection{(Single) Maximum Estimator}

The most straightforward way to approximate $\max_a \mathbb{E} [Q_a]$ is to take the maximum over the sample mean for each $a$, i.e., $\max_a \mathbb{E} [Q_a] \approx \max_a \bar{q}_a $. 
Note that the sample means $\bar{q}_a$ are unbiased estimates of the true means, thus 
$\max_a \bar{q}_a $ is an unbiased estimate for $\mathbb{E} [\max_a \mu_a] = \int_{-\infty}^\infty x f_{\max} (x) dx $, however, it is a biased estimate for $\max_a \mathbb{E} [Q_a]$.

Consider its CDF $F^{\mu}_{\max} = P \{ \max_a \hat{\mu}_a \le x \} =  \Pi_{a} P\{ \mu_a \le x\} = \Pi_a F_a^\mu (x) $, we can write 
\begin{equation}
    \label{eqn:emax}
    \mathbb{E} [\max_a \hat{\mu}_a] = \int_{-\infty}^\infty x \frac{d}{dx} \Pi_a F_a^{\mu} (x) dx = \sum_{a'} \int_{-\infty}^\infty x f_a^\mu (x) \Pi_{a' \ne a} F_a^\mu (x) dx  .
\end{equation}

Comparing equations (\ref{eqn:maxe}) and (\ref{eqn:emax}), clearly $\max_a \mathbb{E} [Q_a] $ and $\mathbb{E} [\max_a \hat{\mu}_a] $ are note equivalent. Moreover, the product term $\Pi_{a' \ne a} F_a^\mu (x)$ in the integral introduces positive bias (since CDFs are monotonically increasing, the sum of their derivatives will be positive, the integral value would be monotonically increasing while more product terms are added).
Therefore, we say that the expected value of the single estimator for the maximum is an overestimation of the maximum expected value.

%Moreover, if (the action space $\mathcal{A}$ is continuous and) we assume $Q_a$ is draw from Gaussian i.i.d., i.e., $Q_a \overset{iid}{\sim} \mathcal{N}(\hat{\mu}, \sigma )$, we can provide both upper and lower bounds for $\mathbb{E}[\max_a Q_a]$. We can extend [\cite{kamath2015bounds}] for deriving $\mathcal{N}(\hat{\mu}, \sigma )$ case, which should not be hard...

\subsection{Double Estimator}
Consider the case that we use two sets of estimators $\hat{\mu}^A = \{\hat{\mu}^A_{a_1}, \cdots, \hat{\mu}^A_{a_{|\mathcal{A}|}} \}$ and $\hat{\mu}^B = \{\hat{\mu}^B_{a_1}, \cdots, \hat{\mu}^B_{a_{|\mathcal{A}|}} \}$, in which each $\hat{\mu}^A_a$ and is estimated from a set of samples independent of the one to estimate $\hat{\mu}^B_a$, i.e., $\hat{\mu}^A_a = \frac{1}{|\Omega^A_a|} \sum_{x\in \Omega^A_a} x$, $\hat{\mu}^B_a = \frac{1}{|\Omega^B_a|} \sum_{x\in \Omega^B_a} x$, and $\Omega^A_a \cap \Omega^B_a = \emptyset$. For all $a$, both $\hat{\mu}^A_a$ and $\hat{\mu}^B_a$ are unbiased estimator for $\mathbb{E} [Q_a]$, assuming all the samples in both sets are independently drawn from the population. That means $\mathbb{E} [\hat{\mu}_a^A] = \mathbb{E} [Q_a]$ for all $a$, including $a^\ast_B = \textrm{argmax}_a \hat{\mu}^B_a $, the action that maximizes the sample mean $\hat{\mu}^B$. Therefore, $\hat{\mu}^A_{a_B^\ast}$ can be used to estimate $\max_a \mathbb{E}[\hat{\mu}^A_a]$ as well as $\max_a \mathbb{E}[Q_a]$, i.e., 
$$\max_a \mathbb{E} [Q_a] = \max_a \mathbb{E}[\hat{\mu}^A_a] \approx \hat{\mu}_{a^\ast_B} ^A.$$

The same argument holds for the opposite way considering the best action over $\Omega^A$ and the sample mean $\hat{\mu}_{a^\ast_A} ^B$. 
% \begin{lem}
% Double estimator gives negative bias.
% \end{lem}
The selection of $a^\ast$ means that all other $a$ gives lower estimation, i.e., $P(a=a^\ast)=\Pi_{a\ne a^\ast}P(\mu_a^A <\mu_{a^\ast}^A)$.
Let $f^A_a$ and $F^A_a$ be the PDF and CDF of $\mu^A_a$, respectively. Then
$$P(a=a^\ast) =  \int_{-\infty}^\infty P(\mu_a^A=x) \Pi_{a'\ne a} P(\mu_A^A < x) dx = \int_{-\infty}^\infty x f_a^A (x) \Pi_{a' \ne a} F_a^A (x) dx  .$$
The expected value of double estimator is a weighted sum of the sample means' expected values in one sample space, weighted by the probability of each sample mean to be the maximum in the other sample space, i.e.,
$$\sum_a P(a=a^\ast) \mathbb{E}[\mu^B_a] = \sum_a \mathbb{E}[\mu^B_a]  \int_{-\infty}^\infty x f_a^A (x) \Pi_{a' \ne a} F_a^A (x) dx . $$

Double estimator gives us negative bias, since the weights $P(a=a^\ast)$ are probabilities, which are positive and sum to 1, the maximum expected value then serves as an upper bound for the weighted sum, as some weights may also be given to variables whose expected value is less than the maximum.

\subsection{Cross Estimator}
We can easily extend the double estimator to a more general case, in which instead of using two sets of estimators, suppose now we have $K$ independent sets of estimators $\hat{\mu^1}, \cdots, \hat{\mu^K}$. We call it the cross estimator. 
The double estimator can be seen as a special case of the more general cross estimator.
Similar argument as analyzing the double estimator can be applied here, for any two estimators $\hat{\mu^i}$ and $\hat{\mu^j}$, as $$\max_a \mathbb{E} [Q_a] = \max_a \mathbb{E}[\hat{\mu}^i_a] \approx \hat{\mu}_{a^\ast_j} ^A.$$ The cross estimator finally uses a convex combination of the $K$ sample means, 
$$\sum_a P(a=a^\ast) \mathbb{E}[\mu^j_a] = \sum_a \mathbb{E}[\mu^j_a]  \int_{-\infty}^\infty x f_a^i (x) \Pi_{a' \ne a} F_a^i (x) dx , $$
thus also underestimates the maximum expected value.
% \begin{thm}
% Cross estimator gives negative bias.
% \end{thm}

\begin{theorem} \cite{van2013estimating}
There does not exist an unbiased estimator for maximum expected values.
\end{theorem}

\section{Convergence in the Limit}
\label{sec:cq_conv}
In this section, we first present a lemma which claims the convergence of SARSA from \cite{singh2000convergence}, and then use it to prove convergence of cross Q-learning. Note that this part heavily borrows the proof of the convergence of double Q-learning \cite{hasselt2010double}, but serves as a more general case.

\begin{lemma} \cite{singh2000convergence} . \label{lemma:converge}
Consider a stochastic process $(\alpha_t, \Delta_t, F_t ), t\ge 0$, where $\alpha_t, \Delta _t$ and $F_t : X \rightarrow \mathbb{R}$ satisfy the equation
$$ \Delta_{t+1} (x) = (1-\alpha_t (x)) \Delta_t(x) + \alpha_t (x) F_t(x), \qquad \textrm{where } x\in X, t = 0, 1, 2, \cdots $$
Let $P_t$ be a sequence of increasing $\sigma$-fields such that $\alpha_0$ and $\Delta_0$ are $P_0$-measureable and $\alpha_t, \Delta_t$ and $F_{t-1}$ are $P_t$-measurable, for $t = 1, 2, \cdots$.

$\Delta_t$ converges to zero with probability one (w.p.1) if the following hold:
\begin{enumerate}
    \item the set $X$ is finite.
    \item $0\le \alpha_t(x) \le 1, \sum_t\alpha_t(x) = \infty$, and $\sum_t \alpha_t^2(x) < \infty$ w.p. 1.
    \item $|| \mathbb{E}[F_t | P_t] || \le \kappa || \Delta_t || + c_t$, where $\kappa \in [0, 1]$ and $c_t$ converges to zero w.p. 1.
    \item $Var(F_t|P_t) \le K(1 + || \Delta_t ||)^2$, where $K$ is a constant.
\end{enumerate}
in which $||\cdot||$ denotes the maximum norm.
\end{lemma}

\begin{theorem} \label{thm:cross_converge}
In a given ergodic MDP, suppose that we have a set of $K$ Q-value functions, $Q^1, Q^2, \cdots, Q^K$, as updated by cross Q-learning, will converge to the optimal value function $Q^\ast$ with probability 1, if the following conditions hold:
\begin{enumerate}
    \item The MDP is finite, i.e., $|\mathcal{S}\times \mathcal{A}| < \infty$.
    \item $\gamma \in [0, 1)$.
    \item The $Q$-values are stored in a lookup table.
    \item Each state-action pair is visited infinitely often. 
    \item Each $Q^k$ receives an infinite number of updates, for all $k = 1, \cdots, K$.
    \item $0\le \alpha_t(s, a) \le 1, \sum_t\alpha_t(s, a) = \infty$, and $\sum_t \alpha_t^2(x) < \infty$ w.p. 1. Moreover, $\alpha_t(s, a) = 0, \forall (s, a) \ne (s_t, a_t)$.
    \item $Var(R(s, a)) < \infty, \forall s, a$
\end{enumerate}
\begin{proof}
Let $k, j \in \{1, \cdots, K \}$ are randomly picked with $k \ne j$.  
Apply Lemma \ref{lemma:converge} by letting $P_t = \{Q_0^1, Q_0^2, \cdots, Q_0^K, s_0, a_0, \alpha_0, r1, s1, \cdots, s_t, a_t \}$, $X = \mathcal{S \times A}$, $\Delta_t = Q^k_t - Q^\ast$, and 
$F_t(s_t, a_t) = r_t + \gamma Q^j_t(s_{t+1}, a^\ast) - Q^\ast_t(s_t, a_t)$, where $a^\ast = \textrm{argmax}_a Q^k (s, a)$. 
The first two conditions of Lemma \ref{lemma:converge} hold immediately from conditions 1 and 6 of Lemma \ref{thm:cross_converge}, respectively. And since condition 7 of Theorem \ref{thm:cross_converge} gives us the bounds for the variance of rewards, the fourth condition of Lemma \ref{lemma:converge} holds.

To show the third condition of Lemma \ref{lemma:converge}, we write
\begin{align*}
    F_t(s_t, a_t) & = r_t + \gamma Q^j_t(s_{t+1}, a^\ast) - Q^\ast_t(s_t, a_t) \\
    & = \left(r_t + \gamma Q^k_t (s_{t+1}, a^\ast) - Q^\ast_t s_t(s_t, a_t) \right) + \gamma \left(Q^j_t(s_{t+1}, a^\ast) - Q^k_t(s_{t+1}, a^\ast)\right) \\
                & = F^Q_t(s_t, a_t) + \gamma c_t %\left(Q^j_t(s_{t+1}, a^\ast) - Q^k_t(s_{t+1}, a^\ast)\right)
\end{align*}
in which we define $F^Q_t = r_t + \gamma Q^k_t (s_{t+1}, a^\ast)$ as the estimated $Q$-value for $(s, a)$ under the standard (single) Q-learning. While the convergence of standard Q-learning in finite MDP is well-known, i.e., $\mathbb{E}[F^Q_t |P_t] \le \gamma ||\Delta_t||$, it suffices to show that $c_t = Q^j_t(s_{t+1}, a^\ast) - Q^k_t(s_{t+1}, a^\ast) \to 0$, so that the condition on the expected contraction of $F_t$ holds.

Let $\Delta^{jk}_t (s_t, a_t) = Q^j_t(s_t, a_t) - Q^k_t(s_t, a_t)$. It is important to note that at each step, the choice of $j,k$ is random, all with equal probability $p_{jk} = p_{jk} = 1/{K\choose 2}$.
Consider the case that $Q^k$ is updated using $Q^j_{t}$ at time $t$, the update of $\Delta^{kj}$ is 
\begin{align*}
    \Delta^{jk}_{t+1} (s_t, a_t) & =  \Delta^{jk}_{t} (s_t, a_t) + \alpha_t(s_t, a_t) \left( r_t + \gamma Q^j_t(s_{t+1}, a^\ast) - Q^k_t(s_t, a_t) \right) \\ 
    & =  \Delta^{jk}_{t} (s_t, a_t) + \alpha_t(s_t, a_t)  F^k_t(s_t, a_t)
\end{align*}
Or, again with probability $p_{kj} = 1/{K\choose 2}$, we use $Q^k$ to update $Q^j$, in this case we have
\begin{align*}
    \Delta^{jk}_{t+1} (s_t, a_t) & =  \Delta^{jk}_{t} (s_t, a_t) - \alpha_t(s_t, a_t) \left( r_t + \gamma Q^k_t(s_{t+1}, a^\ast) - Q^j_t(s_t, a_t) \right) \\ 
    & =  \Delta^{jk}_{t} (s_t, a_t) - \alpha_t(s_t, a_t)  F^j_t(s_t, a_t)
\end{align*}
Otherwise, this particular $(j, k)$ pair is not selected at time $t$, and the update of $\Delta^{kj}$ is then zero. Then
\begin{align*}
    \mathbb{E}\left[ \Delta^{jk}_{t+1} | P_{t+1} \right] %&=   \\
    &= (1 - 2p_{jk}) \mathbb{E}\left[ \Delta^{jk}_{t}  | P_{t} \right] 
\end{align*}
Clearly $\mathbb{E}\left[ \Delta^{jk}_{t+1} | P_t \right] $ converges to 0 since the coefficient on the R.H.S. is less than 1. Therefore we have shown that $c_t \to 0$ since $\Delta^{jk}_t \to 0$ in expectation and $j, k$ are randomly chosen. It in turn ensures condition 3 of Lemma \ref{lemma:converge} holds, which completes our proof.
\end{proof}
\end{theorem}

Finally, we rephrase Theorem \ref{thm:cross_converge} as follows:
\begin{proposition}
Cross estimation converges in the limit, given finite and ergodic MDP.
\end{proposition}
% \end{theorem}

% \end{appendices}
\end{document}